%% file: main.tex
\documentclass[10pt,twocolumn,letterpaper]{article}
\newtheorem{globalcounter}{Theorem}[section] 

\newcounter{ManuCounter}
\newcommand{\manu}[1]{{\small \color{purple} 
\refstepcounter{ManuCounter}\textsf{[ML]$_{\arabic{ManuCounter}}$:{#1}}}}

\newtheorem{definition}[globalcounter]{Definition}

 \newtheorem{observation}[globalcounter]{Observation}
\usepackage{cvpr}              

\input{preamble}

\newcommand{\ourmethod}{E-M3RF\xspace}

\usepackage{siunitx}  
\usepackage{booktabs} 
\usepackage{multirow}
\definecolor{cvprblue}{rgb}{0.21,0.49,0.74}
\usepackage[pagebackref,breaklinks,colorlinks,allcolors=cvprblue]{hyperref}
\usepackage{pgfplots}
\pgfplotsset{compat=newest}

\definecolor{col1}{HTML}{FE6100}
\definecolor{col2}{HTML}{DC267F}
\definecolor{col3}{HTML}{785EF0}
\definecolor{col4}{HTML}{648FFF}
\definecolor{deepblue}{HTML}{003399} 
\definecolor{violet}{HTML}{800080}   
\definecolor{teal}{HTML}{008080}     
\definecolor{amber}{HTML}{FFBF00}  
\definecolor{brown}{HTML}{8B4513}  
\definecolor{blue}{HTML}{1F77B4}  

\def\paperID{13966} 
\def\confName{CVPR}
\def\confYear{2026}

\title{\ourmethod: An Equivariant Multimodal 3D Re-assembly Framework}

\author{%
Adeela Islam$^{1,2}$ \quad Stefano Fiorini$^1$  \quad Manuel Lecha$^1$  \quad Theodore Tsesmelis$^1$ \quad Stuart James$^3$ \\ \quad Pietro Morerio$^1$ \quad Alessio Del Bue$^1$\\
$^1$Fondazione Istituto Italiano di Tecnologia \quad $^2$University of Genova \quad $^3$Durham University \\
\texttt{\{adeela.islam, pietro.morerio\}@iit.it}
}

\begin{document}
\maketitle

\begin{abstract}
%

%
3D reassembly is a fundamental geometric problem, and in recent years it has increasingly been challenged by deep learning methods rather than classical optimization. While learning approaches have shown promising results, most still rely primarily on geometric features to assemble a whole from its parts. As a result, methods struggle when geometry alone is insufficient or ambiguous, for example, for small, eroded, or symmetric fragments. Additionally, solutions do not impose physical constraints that explicitly prevent overlapping assemblies.
To address these limitations, we introduce \ourmethod, an equivariant multimodal 3D reassembly framework that takes as input the point clouds, containing both point positions and colors of fractured fragments, and predicts the
transformations required to reassemble them using $SE(3)$ flow matching.
%
%
Each fragment is represented by both geometric and color features: i) 3D point positions are encoded as rotation-consistent geometric features using a rotation-equivariant encoder, ii) the colors at each 3D point are encoded with a transformer. The two feature sets are then combined to form a multimodal representation.
%
%
We experimented on four datasets: two synthetic datasets, Breaking Bad and Fantastic Breaks, and two real-world cultural heritage datasets, RePAIR and Presious, demonstrating that \ourmethod on the RePAIR dataset reduces rotation error by~\textbf{23.1\%} and translation error by~\textbf{13.2\%}, while Chamfer Distance decreases by~\textbf{18.4\%} compared to competing methods.
\end{abstract}
\section{Introduction}
\label{sec:intro}

Reassembly tasks play a fundamental role in many domains, including reconstructing archaeological 3D artifacts~\cite{Tsesmelis2024ReassemblingTP}, piecing together shredded documents~\cite{paixao2020fast}, molecular docking~\cite{corso2023diffdock}, and solving jigsaw or 3D puzzles~\cite{freeman2006apictorial,Sellan2022BreakingBA}. 
\begin{figure}[t]
\centering
\includegraphics[width=1.\linewidth]{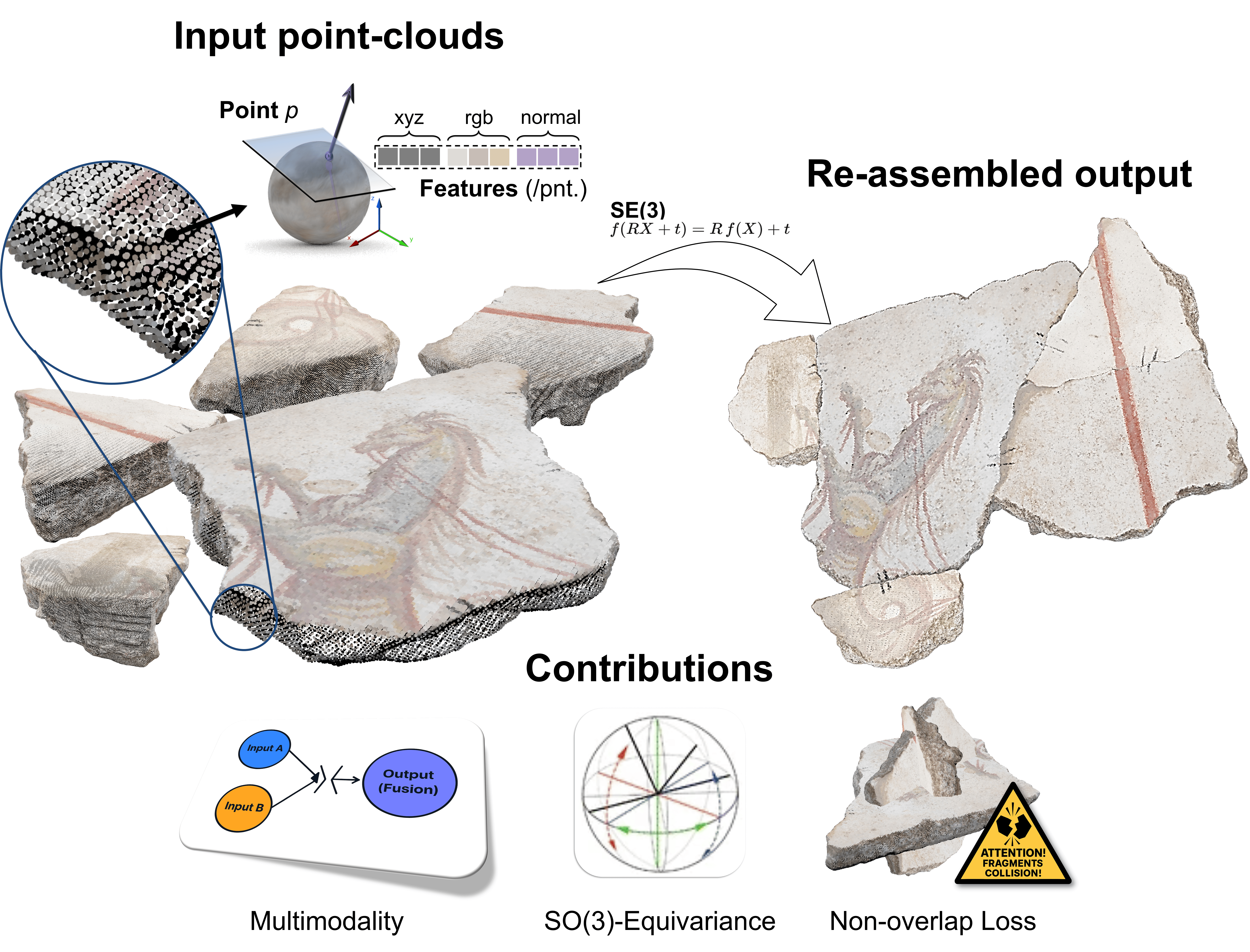}
\vspace{-1.em}
\caption{
\textbf{\ourmethod overview.} Fragments are represented as colored point clouds with per-point features \((x,y,z,\ \mathrm{rgb},\ \mathbf{n})\). 
A multimodal \(\mathrm{SE}(3)\)-equivariant backbone fuses geometry and color and predicts per-fragment rigid transforms \((R,t)\) via flow-based estimation, while a differentiable non-overlap loss enforces physical plausibility. 
The predicted poses reassemble the object, aligning fracture boundaries and color patterns into a coherent result.
}
\label{fig:intro_teaser}
\vspace{-2.0em}
\end{figure}
Accurate reassembly of fractured fragments requires precise estimation of each piece's $SE(3)$ transformation that rebuilds the whole 3D object shape. This problem becomes particularly challenging for symmetric or eroded fragments due to the scarcity of distinctive geometric features. In recent years, 3D fracture reassembly has remained a fundamentally geometric optimization task but has increasingly been approached using deep learning~\cite{Sellan2022BreakingBA, scarpellini2024diffassemble} rather than classical optimization~\cite{zheng2014reassembling, liao20203d}. Learning-based methods have progressed along two complementary trends: improving fragment representations via powerful encoders and pre-training schemes that emphasize fine-scale geometry and surface appearance~\cite{Li2025GARFLG, Islam2025ReassembleNetLK}, and reframing pose estimation as a generative task, where flow- or diffusion-based pipelines iteratively refine fragment alignments~\cite{scarpellini2024diffassemble}.
%
%
%
%
Despite these recent advances, geometry-centric designs have practical limitations: geometric information alone can be ambiguous~\cite{Tsesmelis2024ReassemblingTP}, while real fragments often exhibit appearance details, such as paint, grain, or tool marks. These discriminative cues are ignored by geometry-only pipelines, overlooking an important source of information. Moreover, geometry-only methods often fail to account for the formal group structure of transformations acting on the pieces.
This is compounded by state-of-the-art models' struggle with physical constraints, which often results in intersecting fragments~\cite{scarpellini2024diffassemble, Islam2025ReassembleNetLK}.  

\begin{figure*}[t]
\centering
\includegraphics[width=0.98\linewidth]{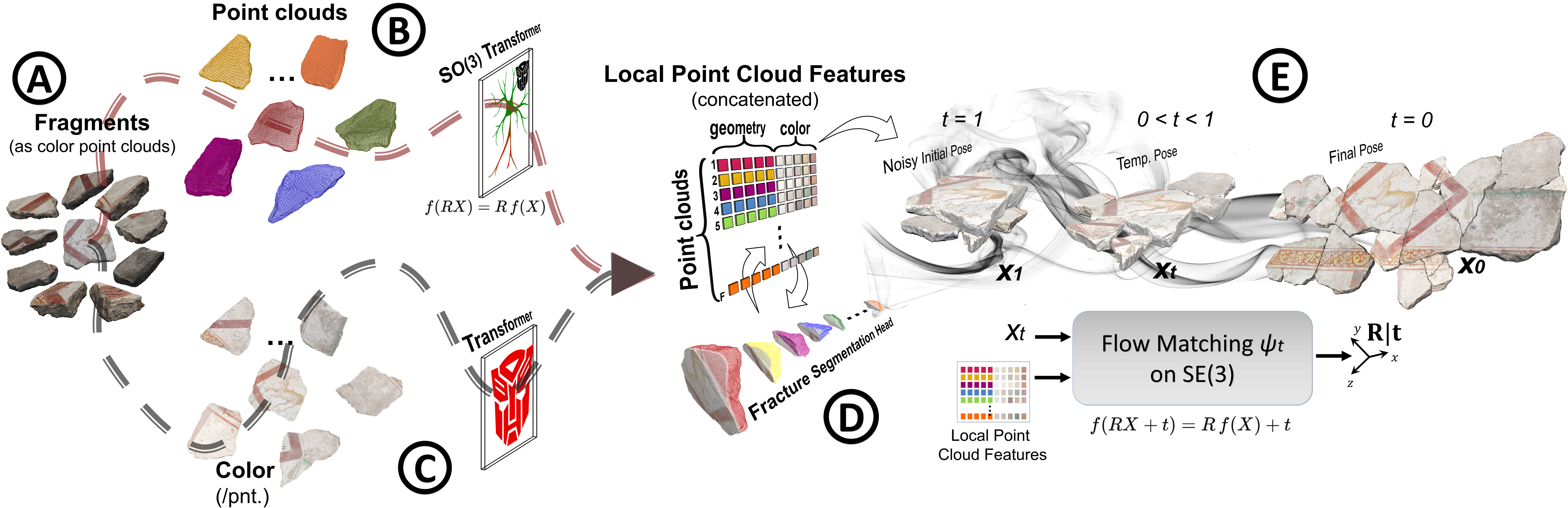}
\vspace{-1.0em}
\caption{
\textbf{\ourmethod's pipeline.} \textbf{\encircle{A}} A set of textured fragments (left) loaded as colored point clouds (w/ per-point RGB). 
Two transformer encoders extract \textbf{\encircle{B}} \textbf{fracture-aware geometric features}, which are geometrically equivariant, since they are processed by an \textbf{\(\mathrm{SO}(3)\)-equivariant Transformer} so representations transform consistently under motions (rotation/ translation) and \textbf{\encircle{C}} \textbf{color features} through a Transformer over per-point colors which extracts dense color descriptors. 
The streams are concatenated into \textbf{\encircle{D}} \textbf{local point-cloud features} (xyz/rgb/normals tokens); the geometry stream is guided by a fracture-segmentation head during pretraining. 
Using these fused features, we perform \textbf{\encircle{E}}  \textbf{flow matching on \(\mathrm{SE}(3)\)}: a time-dependent vector field \(\psi_t\) transports fragment poses from a noisy initialization~\(x_1\) through intermediate states to the assembled configuration~\(x_0\), yielding per-fragment  transforms~\((R,t)\). 
During training, a \textbf{non-overlap loss} penalizes interpenetration to enforce physically plausible assemblies (omitted from the illustration for clarity).
}
\label{fig:teaser}
\vspace{-1.0em}
\end{figure*}

To address the limitations of geometry-only approaches and handle complex fractures, we introduce \ourmethod, a multimodal 3D reassembly framework that takes as input the colored point clouds of fractured fragments and predicts the corresponding $SE(3)$ transformations via $SE(3)$ flow matching~\cite{lipman2023flow, chen2024riemannianflow, joeyse3-fm}, see Fig.~\ref{fig:intro_teaser}. 
The model jointly leverages fragment color and symmetry-aware geometry: an $SO(3)$-equivariant encoder extracts rotation-consistent geometric features from origin-centered point clouds (\ie, no translation involved), while a parallel color branch encodes per-fragment appearance. The resulting features are concatenated into a joint multimodal embedding per fragment point. 
Both rotations and translations are predicted with a transformer-based architecture trained via a flow-matching objective on the $SE(3)$ manifold. 
A differentiable no-overlap loss further enforces physical plausibility by penalizing fragment intersections, as illustrated in Figure \ref{fig:teaser}.

We evaluate \ourmethod on a combination of synthetic and real fracture datasets, including the large-scale synthetic Breaking Bad~\cite{Sellan2022BreakingBA} and Fantastic Breaks~\cite{Lamb2023FantasticBA}, as well as the real-world 
RePAIR~\cite{Tsesmelis2024ReassemblingTP} and Presious~\cite{Theoharis_Papaioannou_2013}.
In all four datasets, we consistently outperform geometry-only baselines, achieving higher part accuracy and lower rotation, translation, and Chamfer Distance errors.
To quantify the impact of our design choices, we perform ablations over each component.


\noindent\textbf{The main contributions of this work are}: 

\begin{itemize}
    \item An end-to-end multimodal fragment representation that using both geometric and color features, producing a joint appearance–geometry embedding that resolves ambiguities and leverages surface patterns when available.
    \item We leverage an inductive bias in the form of a rotation-equivariant geometric backbone. Specifically, $SO(3)$-equivariant layers ensure that features transform consistently under rotations, improving robustness to arbitrary fragment orientations and enhancing generalization.
    \item We enforce physical plausibility via a differentiable no-overlap loss that penalizes fragment intersections.
\end{itemize}

\section{Related Work}
\label{sec:related_work}
We review the state of the art in 3D reassembly and puzzle-solving problems, together with a deep dive into representation encoders commonly used in the solution of 3D tasks.

\subsection{3D Fracture Assembly}

3D puzzle solving and fractured-object reassembly have been approached through a broad range of techniques, spanning classical geometric registration, learning-based pose estimation, and more recent generative formulations. Classical pipelines rely on hand-crafted local descriptors, correspondence hypothesis testing, and ICP-style refinement~\cite{Wang2022BatchbasedMR, Niu2022AnAC}. These methods provide interpretable matching criteria grounded in geometry and remain effective for clean, low-ambiguity fragments. 
Learning-based methods shift to data-driven feature extraction and direct pose prediction. Recent works learn per-point or per-fragment embeddings and estimate 6-DoF alignments using regression or transformer-based architectures~\cite{Li2025GARFLG, Lu2023JigsawLT, scarpellini2024diffassemble, Wang2024PuzzleFusionA3}, often supported by large-scale synthetic fracture corpora such as Breaking Bad~\cite{Sellan2022BreakingBA}. 
Generative approaches, including diffusion or flow-based estimators~\cite{Li2025GARFLG, scarpellini2024diffassemble, Wang2024PuzzleFusionA3}, treat assembly as sampling or transporting distributions over $SE(3)$, enabling iterative refinement and uncertainty-aware reasoning. These methods represent a growing direction for globally coherent reconstruction. 

Although real-world fragments contain useful color cues, existing learning-based reassembly methods rely only on geometric information. \ourmethod addresses this limitation by fusing geometric and color features into a unified multimodal representation, enabling more discriminative fragment descriptors and more reliable alignment.

\subsection{Representation (Encoder)}

PointNet~\cite{qi2017pointnet} introduced a shared-MLP encoder with global feature aggregation, proving that deep learning on point sets is effective for 3D tasks. 
PointNet++~\cite{qi2017pointnetplusplus} extended PointNet with hierarchical neighborhood grouping, allowing the network to capture local geometric features across multiple scales.
Subsequent methods strengthened local feature modeling through convolutional and graph-based operators. 
EdgeConv~\cite{wang2019dynamic} captures point-to-point relationships via dynamic graphs, while KPConv~\cite{thomas2019kpconv} employs kernel-point convolutions as a continuous 3D filtering mechanism, providing strong geometric inductive bias.
Transformer-based encoders have recently gained prominence in 3D vision. 
Point Transformer~\cite{zhao2021pointtransformer} and PCT~\cite{guo2021pct} use self-attention to capture long-range point dependencies, while hierarchical variants like Swin3D~\cite{fan2022swin3d} apply windowed attention to 3D grids.
However, these models are not inherently rotation-aware and do not ensure consistent features under arbitrary $SO(3)$ transformations.
To address this issue, Vector Neurons~\cite{deng2021vector} lift point features from scalars to 3D vectors, allowing rotations to act via matrix multiplication. 
Building on this representation, the VN-Transformer~\cite{deng2023vntransformer} extends self-attention to vector features, yielding a rotation-equivariant attention mechanism that preserves $SO(3)$ consistency throughout the network and enables it to reason about 3D geometry while respecting rotational symmetries.

To effectively address reassembly scenarios where fragments undergo arbitrary rotations, we leverage this rotation-equivariant inductive bias so that the backbone inherits $SO(3)$ transformations directly into the learned features, providing \ourmethod with representations that remain geometrically consistent under all input orientations.

\section{Problem Statement}~\label{sec:problem_statement}
We formalize the geometric object reassembly problem given a finite collection of disjoint fragments.

\noindent\textbf{Fragments.} Let $P=\{p_i\}_{i=1}^N$ be a finite set of points. A fragment is defined as the tuple
\begin{equation}\label{eq:fragment_data}
\mathcal{F} = \bigl(P,\;\mathcal{X},\;\{\mathcal{G}_k\}_{k=1}^{K_{\mathrm{geo}}},\;\{\mathcal{A}_k\}_{k=1}^{K_{\mathrm{app}}}\bigr).
\end{equation}
The map~$\mathcal{X}:P\to\mathbb{R}^3_{\mathrm{xyz}}$ assigns spatial coordinates, \ie the 3D coordinates for each point. Each geometric attribute~$\mathcal{G}_k:P\to\mathbb{R}^3$ transforms under the standard vector representation of~$SE(3)$ (\eg, normals), while each appearance attribute~$\mathcal{A}_k:P\to\mathbb{R}^{d_k}$ is invariant under rotations (\eg, color). In general, there can be~$K_{\mathrm{geo}}~\in \mathbb{N}_0$ geometric attributes and~$K_{\mathrm{app}}~\in~\mathbb{N}_0$ appearance attributes.
Fixing an ordering of the points induces two structured feature spaces:
\begin{equation} \label{eq:geo_app_spaces}
V_{\mathrm{geo}} = \mathbb{R}^N \otimes \mathbb{R}^{K_{\mathrm{geo}}+1} \otimes \mathbb{R}^3, \quad V_{\mathrm{app}} = \mathbb{R}^N \otimes \mathbb{R}^{d_{\mathrm{app}} },
\end{equation}
where the $K_{\mathrm{geo}}+1$ geometric channels include $\mathcal{X}$ as a dedicated vector channel, and $d_{\mathrm{app}} = \sum_{k=1}^{K_{\mathrm{app}}} d_k$ is the total dimensionality of all appearance channels. The full fragment representation lies in
\begin{equation}\label{eq:fragment_space}
V_F = V_{\mathrm{geo}} \oplus V_{\mathrm{app}},    
\end{equation}
and we write $F\in V_F$ for the feature tensor associated with a fragment. We denote its positional component by~$F_{\mathrm{xyz}}~\in~\mathbb{R}^{N}\otimes\mathbb{R}^{3}$.

\noindent \textbf{Unassembled Object.} 
An unassembled object is represented as a collection of $M$ fragments $\mathcal{F}$, each recentered so that its centroid lies at the origin:
\[
\hat{\mathcal{O}} =  \Big\{\hat{F}_i \in V_F \;\Big|\; \frac{1}{N_i}\,\mathbf{1}_{N_i}^\top \hat{F}_{\mathrm{xyz},i} = \mathbf{0} \in \mathbb{R}^3\Big\}_{i=1}^M,
\]
where $\mathbf{1}_{N_i}$ is the all-ones vector of length $N_i$.

\noindent\textbf{Assembled Object.} We assume the existence of a canonical assembled configuration of the object. Each centered fragment $\hat{F}_i$ is placed in this configuration by a rigid transformation $g_i \in SE(3)$, acting exclusively on its positional channel. The assembled object is then denoted by 
\[
\mathcal{O} = \{\, F_i = g_i \cdot \hat{F}_i\,\}_{i=1}^M.
\]

\noindent \textbf{Object Reassembly.} Given any unassembled object $\hat{\mathcal{O}}=\{\hat{F}_i\}_{i=1}^M$, the goal is to recover rigid transformations $\tilde{g}_i \in SE(3)$ that approximate the unknown ground-truth $g_i$.
We therefore seek a single model
\[
\Phi : \hat{\mathcal{O}} \to SE(3)^M, \qquad
\Phi(\hat{\mathcal{O}}) \approx \{\, g_i \,\}_{i=1}^M,
\]
that operates on arbitrary unassembled objects and whose predictions transform and assemble the fragments:
\[
\Phi(\hat{\mathcal{O}})\,\cdot\,\hat{\mathcal{O}} \;\approx\; \mathcal{O}.
\]

\section{Methodology}
\label{sec:method}

\begin{figure}[t]
\centering
\includegraphics[width=\linewidth]{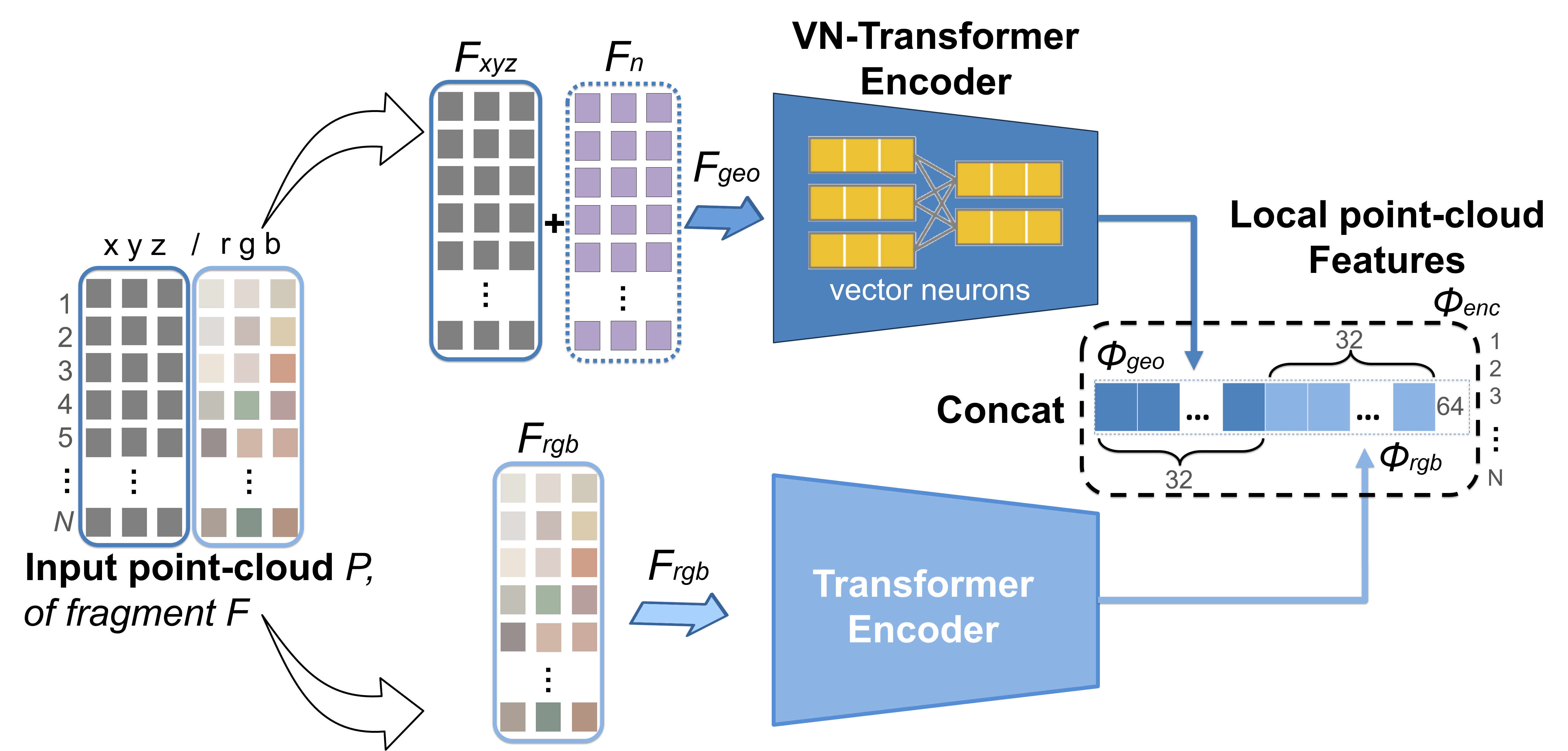}
\vspace{-1.0em}
\caption{
\textbf{Geometry/color feature extraction and fusion.}
The input point cloud \(P\) (per-point \(xyz\), \(rgb\) and normals \(F_{\text{n}}\)), of fragment $\mathcal{F}$ is split into \(F_{\text{geo}}\) and \(F_{\text{rgb}}\).
A \textbf{VN-Transformer encoder} ($SO(3)$-equivariant via vector neurons) produces geometric features \(\Phi_{\text{geo}}\), while a standard \textbf{Transformer encoder} produces color features \(\Phi_{\text{rgb}}\). The features are concatenated per point to yield \(\Phi_{\text{enc}} = [\Phi_{\text{geo}} \,\Vert\, \Phi_{\text{rgb}}\,]\), which serve as the local point-cloud representation for downstream pose estimation.
}
\label{fig:architecture}
\vspace{-1.0em}
\end{figure}

We identify two capabilities required for object reassembly: (1) a representation expressive enough to capture the geometric and appearance cues that govern how fragments fit together, and (2) a mechanism that uses this representation to predict the rigid transformations realizing the correct assembly. 
Having formalized the task in Section~\ref{sec:problem_statement}, our methodology is built to satisfy these requirements in a principled, symmetry–aware way, see Fig.~\ref{fig:architecture}. 
We first learn fragment–level descriptors using an $S_N \times SO(3)$–equivariant encoder, ensuring that the learned representations respect the intrinsic permutation and rotational symmetries of the fragment space (details in the suppl. material).
These descriptors condition a geometric generative model---a conditional $SE(3)$ Riemannian flow-matching network---which predicts a coherent set of rigid transformations aligning fragments into their assembled configuration. 
The full pipeline couples equivariant representation learning with flow-based geometric reasoning, yielding a unified approach that maps any unassembled object $\hat{\mathcal{O}}$ to an assembled one.

\subsection{Encoding Fragments}
\label{subsec:encoding-fragments}

\noindent \textbf{Colored Fragments.}  We consider fragments whose per-point attributes consist of positions, surface normals, and RGB color. Following the formulation in Eq.~\ref{eq:fragment_data} of Sec.~\ref{sec:problem_statement} , each fragment is represented as
\[
\mathcal{F} = (P,\, \mathcal{X},\, \{\mathcal{N}\},\, \{\mathcal{C}\}), \quad \mathcal{N}:P\to\mathbb{R}^3,\quad \mathcal{C}:P\to\mathbb{R}^3,
\]
where $\mathcal{X}:P\!\to\!\mathbb{R}^3$ gives point coordinates, $\mathcal{N}$ gives surface normals, and $\mathcal{C}$ gives per-point color. Hence, the geometric and appearance feature spaces in Eq.~\ref{eq:geo_app_spaces} specialize to
\[
V_{\mathrm{geo}} = \mathbb{R}^N \otimes \mathbb{R}^2 \otimes \mathbb{R}^3,
\qquad
V_{\mathrm{rgb}} = \mathbb{R}^N \otimes \mathbb{R} \otimes \mathbb{R}^3,
\]
where the two geometric vector channels in $V_{\mathrm{geo}}$ correspond to the coordinates $\mathcal{X}$ and the normals $\mathcal{N}$. Each fragment is represented by an element $F \in V_F$ in the joint space of Eq.~\ref{eq:fragment_space}, with canonical projections: $F_{\mathrm{geo}}\in V_{\mathrm{geo}}$ collecting the geometric vector channels; $F_{\mathrm{xyz}}\in \mathbb{R}^N\otimes\mathbb{R}^3$ giving the point coordinates; $F_{\mathrm{n}}\in \mathbb{R}^N\otimes\mathbb{R}^3$ giving the normals; and $F_{\mathrm{rgb}}\in V_{\mathrm{rgb}}$ giving the per-point color vectors.

\noindent \textbf{Fragment Symmetries.} Since each fragment in $\hat{\mathcal{O}}$ is centered at the origin, its continuous geometric symmetry reduces to global rotations
$SO(3)\subset SE(3)$. In addition, the indexing of points induces a discrete permutation symmetry described by the symmetric group $S_N$. The resulting symmetry group acting on geometric features is therefore
\begin{equation*}\label{eq:group}
\mathbf{G} \coloneqq S_N \times SO(3),
\end{equation*}
where $S_N$ permutes point indices and $SO(3)$ acts on every geometric
$\mathbb{R}^3$ channel. For $\sigma\in S_N$, let $P_\sigma$ be the associated permutation matrix, and let $I_2$ denote the identity on the two geometric channels. The induced action on $F_{\mathrm{geo}}\in V_{\mathrm{geo}}$ is
\begin{equation}\label{eq:group_action}
(\sigma,R)\cdot F_{\mathrm{geo}} = (P_\sigma \otimes I_2 \otimes R)\, F_{\mathrm{geo}},  \quad (\sigma,R)\in\mathbf{G},
\end{equation}
where $\otimes$ is the Kronecker product implementing the tensor-product representation. In pointwise notation, for~$i\in~\{1,\dots,N\}$,
\[
((\sigma,R)\cdot F_{\mathrm{geo}})_i
=
\begin{bmatrix}
R\, (F_{\mathrm{xyz}})_{\sigma^{-1}(i)} \\[3pt]
R\, (F_{\mathrm{n}})_{\sigma^{-1}(i)}
\end{bmatrix}
\in \mathbb{R}^{2} \otimes \mathbb{R}^3,
\]
where the first component corresponds to permuted and rotated coordinates and the second to permuted and rotated normals. 
For color features $F_{\mathrm{rgb}} \in V_{\mathrm{rgb}}$, the $SO(3)$ factor acts trivially. Writing the induced action using the same tensor-product structure,
\begin{equation}\label{eq:rgb_action}
(\sigma,R)\cdot F_{\mathrm{rgb}} \;=\; (P_\sigma \otimes I_1 \otimes I_3) \,F_{\mathrm{rgb}}, \quad (\sigma,R)\in\mathbf{G},
\end{equation}
reflecting that RGB vectors are rotation-invariant and only their pointwise ordering is affected.

\noindent \textbf{Geometric Encoder $\Phi_{\mathrm{geo}}$.} The geometric branch implements a $\mathbf{G}$-equivariant map using a VN-Transformer~\cite{deng2023vntransformer},
\[
\Phi_{\mathrm{geo}}: V_{\mathrm{geo}} \to  \mathbb{R}^{N} \otimes \mathbb{R}^{C_{\mathrm{geo}}} \otimes \mathbb{R}^{3},
\]
which outputs $C_{\mathrm{geo}}$ equivariant vector channels, each with values in $\mathbb{R}^3$. The $\mathbf{G}$-action on this output space is obtained by lifting the action in Eq.~\ref{eq:group_action} from the two input geometric channels to the $C_{\mathrm{geo}}$ learned channels. As a result, $\Phi_{\mathrm{geo}}$ satisfies
\[
\Phi_{\mathrm{geo}}\!\bigl((P_\sigma \otimes I_{2} \otimes R)\,
     F_{\mathrm{geo}}\bigr)
=
(P_\sigma \otimes I_{C_{\mathrm{geo}}} \otimes R)\,
     \Phi_{\mathrm{geo}}(F_{\mathrm{geo}}),
\]
for all $(\sigma,R)\in\mathbf{G}$. Here $I_{C_{\mathrm{geo}}}$ denotes the identity in $\mathbb{R}^{C_{\mathrm{geo}}}$.

\noindent \textbf{Color Encoder $\Phi_{\mathrm{rgb}}$.} Color features are encoded by a Transformer
\[
\Phi_{\mathrm{rgb}} : V_{\mathrm{rgb}} \to \mathbb{R}^{N} \otimes \mathbb{R}^{C_{\mathrm{rgb}}} \otimes \mathbb{R}^{3},
\]
which operates on rotation–invariant channels. Consequently, the only nontrivial component of the $\mathbf{G}$-action is the permutation action of $S_N$. Thus, the $SO(3)$ factor of
$\mathbf{G}$ acts trivially on $V_{\mathrm{rgb}}$, and the only nontrivial group
action arises from permutations of point indices.   Lifting the permutation action of Eq.~\ref{eq:rgb_action} to the output space, $\Phi_{\mathrm{rgb}}$ satisfies
\[
\Phi_{\mathrm{rgb}}( (P_\sigma \otimes I_{1} \otimes I_3) F_{\mathrm{rgb}}) = (P_\sigma \otimes I_{C_{\mathrm{rgb
}}} \otimes I_3) \, \Phi_{\mathrm{rgb}}(F_{\mathrm{rgb}}), 
\]
for all $\sigma\in S_N$.

\noindent \textbf{Fused Encoder $\Phi_{\mathrm{enc}}$.} Let $C = C_{\mathrm{geo}} + C_{\mathrm{rgb}}$. We define the fused encoder
\[
\Phi_{\mathrm{enc}} : V_F \longrightarrow  \mathbb{R}^{N} \otimes \mathbb{R}^{C} \otimes \mathbb{R}^{3},
\]
by
\[
\Phi_{\mathrm{enc}}(F) = \bigl[ \Phi_{\mathrm{geo}}(F_{\mathrm{geo}}) \;\Vert\; \Phi_{\mathrm{rgb}}(F_{\mathrm{rgb}}) \bigr],
\]
where $\Vert$ denotes concatenation along the feature–channel axis. For each point $i$, the geometric block $\Phi_{\mathrm{geo}}(F_{\mathrm{geo}})_i \in \mathbb{R}^{C_{\mathrm{geo}}}\otimes\mathbb{R}^3$
and the color block $\Phi_{\mathrm{rgb}}(F_{\mathrm{rgb}})_i \in \mathbb{R}^{C_{\mathrm{rgb}}}$ are merged into a single $C$-channel tensor. Since both branches respect the $\mathbf{G}$-action on their respective representation spaces, the fused encoder $\Phi_{\mathrm{enc}}$ is $\mathbf{G}$-equivariant by construction.

\subsection{Fracture-Boundary Segmentation Pretraining}

The encoder is pretrained via an auxiliary fracture-boundary segmentation task, which encourages the learned fragment representations to capture the geometric signatures of fracture interfaces. For each fragment $\hat{F}_i$, the encoder produces the pointwise  embedding 
\begin{equation}\label{eq:encoded_fragment}
H_i = \Phi_{\mathrm{enc}}(\hat{F}_i) \in \mathbb{R}^{N_i} \otimes \mathbb{R}^{C} \otimes \mathbb{R}^{3}.
\end{equation}
A lightweight segmentation head 
\[
\Psi_{\mathrm{seg}}: H_i \longrightarrow \hat{y}_i,  \qquad \hat{y}_i \in [0,1]^{N_i},
\]
predicts per-point boundary probabilities, supervised by binary ground-truth masks. The  pretraining objective optimizes the composition $\Psi_{\mathrm{seg}} \circ \Phi_{\mathrm{enc}}$ to correctly identify fracture boundaries, forcing $\Phi_{\mathrm{enc}}$ to learn geometry–appearance interactions characteristic of fracture morphology. This procedure yields symmetry-preserving embeddings that serve as inputs to the subsequent flow-matching model on $SE(3)$.

\subsection{Fragment Assembly}

Given the encoded fragment features (Eq.\ref{eq:encoded_fragment}),~$H_i =~\Phi_{\mathrm{enc}}(\hat{F}_i)$, we enrich each fragment with geometric priors using a NeRF-style positional encoding. Specifically, we concatenate multi-frequency sinusoidal embeddings~$\mathrm{PE}(\cdot)$~\cite{tancik2020fourier} of point cloud coordinates~$\hat{F}_{\mathrm{xyz}}$, normals~$\hat{F}_{\mathrm{n}}$, and scale information~$s_i~\in~\mathbb{R}$ along the channel dimension. Let~$\mathcal{H} \coloneqq \mathbb{R}^{N_\star} \otimes \mathbb{R}^{C_{\mathrm{enc}}} \otimes \mathbb{R}^{3},$ where~$N_\star$ is a fixed maximum number of points. The enriched fragment representation becomes:
\[
H_i :=f_{shape}\left(\big[H_i \, \Vert \, \mathrm{PE}((\hat{F}_{\mathrm{xyz}})_i) \, \Vert \, \mathrm{PE}((\hat{F}_{\mathrm{n}})_i) \, \Vert \, \mathrm{PE}(s_i)\big]\right),  
\]
where $f_{shape}(\cdot)$ is the shape embedding function. This provide high-frequency geometric cues that improve pose estimation. We additionally introduce a learned \emph{anchor token} that designates the reference fragment around which all poses are predicted.  
Conditioned on $H_i$, we define a conditional Riemannian flow-matching model on the Lie group $SE(3)$ which learns a continuous velocity field whose integration yields the rigid transformation $g_i = (R_i,\beta_i) \in SE(3)$ required to place the fragment $i$ in its correct location within the assembled object. Formally, the model parametrizes a time-dependent velocity field
\[
v : SE(3) \times [0,1] \times \mathcal{H}  \longrightarrow \mathfrak{se}(3) = \mathfrak{so}(3) \times \mathbb{R}^3,
\]
which associates to each element $g\in SE(3)$, time $t\in[0,1]$, and conditioning features $H_i\in\mathcal{H}$ a tangent vector $v(g,t;H_i)\in\mathfrak{se}(3)$. During training, the flow matching objective aligns this vector field with the target probability path interpolating between an initial distribution
\[
p_0(g) = \mathcal{U}(SO(3)) \otimes \mathcal{N}(\mathbf{0}, I_3)
\]
and the empirical distribution of assembled configurations. 
At inference time, the learned vector field is integrated over the unit interval as
\[
\dot{g}_i^{\,t} = v\big(g_i^{\,t},\, t;\, H_i\big), \qquad  g_i^{\,0} = (I,\mathbf{0}),
\]
yielding the final transformations $g_i^1$ reassembling the object:
\[
\mathcal{O} = \big\{\, g_i^1 \cdot \hat{F}_i \,\big\}_{i=1}^M.
\]
This formulation defines a continuous, symmetry-preserving flow on $SE(3)$ that transports fragments from their initial centered configuration to globally consistent positions in the assembled object. In detail, the conditional trajectory between an initial pose $g_i^0$ and the target pose $g_i^1$ follows a geodesic interpolation in rotation and a linear interpolation in translation:
$$
R_i^{t} = \exp_{R_i^{0}}\!\big(t\,\log_{R_i^{0}}(R_i^{1})\big), 
\quad
\beta_i^{t} = (1-t)\,\beta_i^{0} + t\,\beta_i^{1}.
$$
Accordingly, let us denote the rotational and translational velocity fields as
$$
v_R^{(i)}(g^t_i,t) \in \mathfrak{so}(3), 
\qquad 
v_\beta^{(i)}(g^t_i,t) \in \mathbb{R}^3.
$$
Following the flow-matching formulation, these predicted velocities are trained to match the target geodesic residuals scaled by $(1-t)^{-1}$. Hence, the objective is expressed as:
\begin{align}
\mathcal{L}_\text{flow}
=& 
\mathbb{E}_{t,\;p_1(g_i^1),\;p_t(g_i^t \mid g^1)}\!\Bigg[
\frac{1}{N} \sum_{i=1}^N 
\Bigg(
\big\| v_\beta^{(i)}(g^t_i,t)
- \tfrac{\beta^1_{i} - \beta^t_{i}}{1-t} \big\|_2^2
\nonumber\\
&+ 
\lambda\, 
\big\| v_R^{(i)}(g^t,t)
- \tfrac{1}{1-t}\,\log_{R^t_{i}}\left(R^1_{i}\right) \big\|_2^2
\Bigg)
\Bigg],
\end{align}
where $\log_{R^t_{i}}\left(R^1_{i}\right) \in \mathfrak{so}(3) \cong \mathbb{R}^3$ denotes the Lie-algebra residual from $R^t_{i}$ to $R^1_{i}$, and $\lambda$ balances rotational and translational terms. This formulation ensures that predicted fragment transformations follow the correct geodesic trajectories on $SE(3)$, maintaining smooth and consistent rotational and translational flows.








%

\subsection{No-Overlap Loss} 
To prevent implausible fragment intersections, we introduce a differentiable {no-overlap loss}. 
For each pair of fragments~$(i,j)$, we compute soft occupancy masks~$M_i(x), M_j(x) \in [0,1]$ denote the soft occupancy values of fragments $i$ and $j$ at volume cell (grid location) $x$ in a regular volumetric grid. We define the pairwise IoU:
\begin{equation}
    \mathrm{IoU}_{ij} = \frac{\sum_x \min\left(M_i(x), M_j(x)\right)}{\sum_x \max \left(M_i(x), M_j(x) \right) + \varepsilon},
\end{equation}
where $\varepsilon>0$ ensures numerical stability. The no-overlap loss aggregates these pairwise terms:
\begin{equation}
    \mathcal{L}_\text{no-overlap} = \frac{1}{|P|} \sum_{(i,j)\in P} \mathrm{IoU}_{ij}.
\end{equation}
Minimizing $\mathcal{L}_\text{no-overlap}$ enforces non-overlap between fragments, improving visual quality and physical plausibility.

The full loss combines the flow-matching term and the no-overlap penalty:
\begin{equation}
    \mathcal{L} = \mathcal{L}_\text{flow} + \alpha \, \mathcal{L}_\text{no-overlap},
\end{equation}
where $\alpha$ balances pose accuracy with overlap prevention. 
Together, the two-stage training and differentiable no-overlap loss enable the network to produce $SE(3)$ predictions that are both accurate and physically consistent.

\section{Experimental Evaluation}
\label{sec:eval}

We evaluate our method on four different dataset. 
The datasets, evaluation metrics, and competing methods are described in Section~\ref{subsec:dataset}, while Section~\ref{subsec:result} reports the performance of all methods along with some qualitative results.
In Section~\ref{subsec:ablation}, we present an ablation study examining how \ourmethod generalizes across different objects and datasets\footnote{Further ablations—including point sampling and runtime—appear in the suppl. material.}. 

\subsection{Dataset and Evaluation Metrics}\label{subsec:dataset}

\textbf{Datasets.} Similarly to the evaluation protocol described in~\cite{Li2025GARFLG}, we use a combination of large-scale synthetic and real scanned datasets to probe complementary aspects of 3D assembly. 
For the colorless (geometry-only) setting, we us~(i) Breaking Bad~\cite{Sellan2022BreakingBA}, a large-scale synthetic dataset of over a million fractured instances derived from around 10k base shapes (PartNet/Thingi10k), which provides clean ground-truth poses and diverse geometry for controlled ablations and large-scale pretraining; and (ii) Fantastic Breaks~\cite{Lamb2023FantasticBA}, offering paired scans of intact and broken real objects that capture realistic micro-fracture morphology and scanning artifacts for high-fidelity geometric evaluation.
For the colored fragments (multimodal setting), we evaluate on RePAIR~\cite{Tsesmelis2024ReassemblingTP} and the Presious~\cite{Theoharis_Papaioannou_2013} datasets.
The former is a collection of archaeological fresco fragments with high-resolution RGB imagery and 3D geometry, and the latter is a set of several 3D cultural heritage fragment groups. 
Both are well suited to assessing color-based cues and multimodal fusion challenges, like erosion and missing-parts.
Due to the size of the Presious dataset, \ie only six sets, we use it only for testing the generalization accuracy of \ourmethod.
We also reference FRACTURA\footnote{Not publicly available at submission; excluded from our experiments.}~\cite{Li2025GARFLG}, curated with archaeologists includes different types of fragments 
for evaluating generalization to realistic breakage.  
Accordingly, we report the all “with color” results only on RePAIR and Presious datasets, while geometry-only ablations and generalization studies are conducted on all aforementioned datasets.
Following~\cite{Wang2024PuzzleFusionA3,Li2025GARFLG}, we sample 5k points per object, ensuring uniform per-fragment density.

\noindent\textbf{Metrics.} We report four measures: \textbf{RMSE ($\mathcal R^\circ$)} is the root-mean-square rotation error (in degrees); \textbf{RMSE ($\mathcal T_{\rm mm}$)} is the root-mean-square translation error (in millimeters); \textbf{PA} (Part Accuracy) is the percentage of fragments whose per-fragment Chamfer Distance to the ground-truth placement falls below 0.01\%; and \textbf{CD} is the Chamfer Distance between the assembled object point cloud and the ground-truth assembly. 
We keep the same definitions and threshold as in~\cite{Li2025GARFLG} to ensure a fair comparison across datasets and baselines.
The best performance metric is highlighted in \textbf{bold}, while the second-best competitor is \underline{underlined}.


\subsection{Results}\label{subsec:result}

Table~\ref{tab:3d_results_repair} reports a head-to-head comparison on the color-enabled RePAIR dataset. 
The full \ourmethod model (last row) achieves the best score on all four metrics. 
It reduces rotation and translation errors by 23.1\% and 13.2\% respectively, and Chamfer Distance by 18.4\%, while increasing Part Accuracy by 15.3\% relative to the second best competitor model, GARF~\cite{Li2025GARFLG}. 
The margins over the rest geometric based methods are substantially larger across all metrics, underscoring that a geometry-only pipeline leaves disambiguating information untapped on colored fragments.

\begin{table}[htb!]
  \centering
\setlength{\tabcolsep}{2pt} 
\resizebox{\columnwidth}{!}{%
\begin{tabular}{lrrrr}  
 \toprule
  Method & \multicolumn{1}{c}{\textbf{RMSE ($\mathcal{R}^\circ$)↓}} 
  & \multicolumn{1}{c}{\textbf{RMSE ($\mathcal{T}_{mm}$)↓}} 
  & \multicolumn{1}{c}{\textbf{PA (\%)$\uparrow$ }}
  & \multicolumn{1}{c}{\textbf{CD}↓} \\
    \midrule
  DiffAssemble~\cite{scarpellini2024diffassemble} & 69.54 & 67.96 & 17.92 & 4.18 \\
  PMTR~\cite{Lee20243DGS} & 76.72 & 71.40 & 12.24 & 5.23 \\
  PF++~\cite{Wang2024PuzzleFusionA3} & 67.75 & 70.49 & 20.82 & 18.97 \\
  GARF~\cite{Li2025GARFLG} &\underline{49.31} & \underline{32.19} & \underline{31.14} & 
  \underline{2.66}\\
    \textbf{\ourmethod} (w/o color) & 40.17 & 29.01 & 33.71 & 2.98\\ \hline
  \textbf{\ourmethod} (w/o non-overl.) loss & 45.81 & 25.01 & 32.91 &  3.83\\
  \textbf{\ourmethod} (w/ rotation-inv.) & 43.27 & 41.60 & 32.50 & 3.71 \\
  \textbf{\ourmethod} (w/o rotation-equiv.) & 43.22 & 29.28 & 32.26 &  3.07\\

  \hdashline
  \textbf{\ourmethod} & \textbf{37.91} & \textbf{27.93} & \textbf{35.91} &  \textbf{2.17}\\
\bottomrule

 \end{tabular}}
      \caption{Quantitative results on RePAIR dataset. See Figure~\ref{fig:qualitative} for corresponding qualitative results.}
  \label{tab:3d_results_repair}
\end{table}

\begin{figure*}[t]
\centering
\includegraphics[width=.98\linewidth]{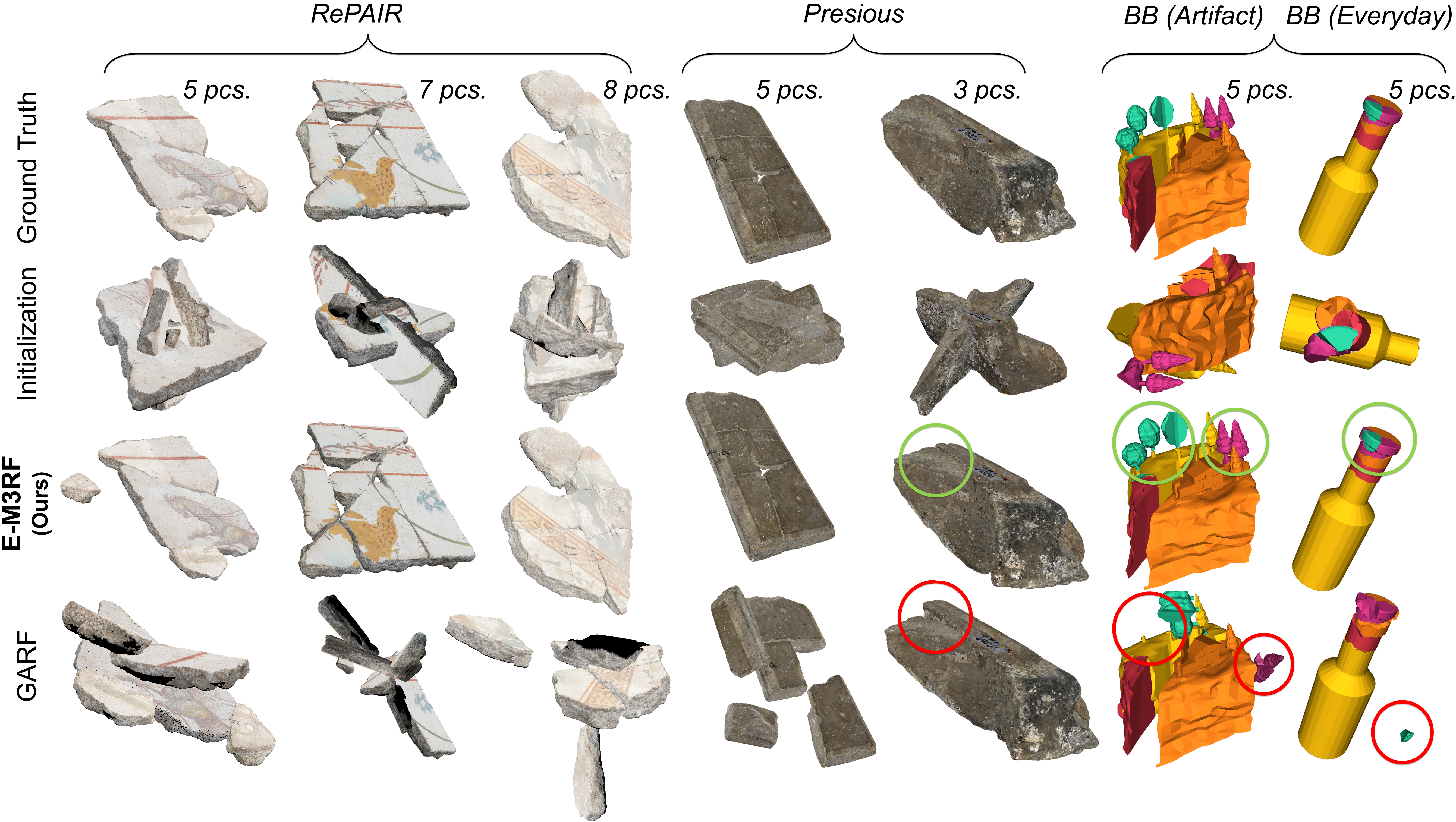}
\caption{
\textbf{Qualitative Comparisons on the RePAIR, Presious and Breaking Bad (BB).}  \ourmethod consistently produces more accurate re-assemblies.
Especially on the Presious scenes, is demonstrating strong generalization to unseen
object.
Green circles denote fine, ambiguous contact regions correctly recovered by our method.
Additional results are available in the supplementary material.
}
\label{fig:qualitative}
\vspace{-1.0em}
\end{figure*}

\ourmethod exploits its superiority mainly via the dense per-point color features fused with fracture-aware geometry, while its $SO(3)$-equivariant backbone makes the learned features transform consistently under pose, improving rotation estimates in particular. 
The non-overlap loss further suppresses physically implausible interpenetrations, reflected in the lower CD despite tighter placement (higher PA).
This is corroborated by the ablations that we have conducted, where we examine the affect of each of our contributions to the overall pipeline:

\begin{itemize}
    \item \textbf{w/o color:} Removing the color branch causes \ourmethod to rely solely on geometric features. 
    PA decreases by 2.2 percentage points (pp), CD increases by 37\%, rotation error increases by 6\%, and translation error increases by 3.9\% in comparison to the full model.
    These results indicate that color provides complementary cues to geometry, helping the model distinguish similar surfaces and resolve ambiguities geometry alone cannot.
    \item \textbf{w/o non-overlap loss:} Removing the non-overlap loss causes the largest increase in CD (76\%) and a decrease in PA by 3.0 pp. Interestingly, translation RMSE decreases (27.93→25.01 mm). 
    However, this improvement is misleading: without the non-overlap loss, fragments interpenetrate, yielding closer centroids but worse CD and PA. 
    These results confirm that the non-overlap loss is essential for plausible placements and accurate surface alignment, as it can be clearly seen in Fig.~\ref{fig:qualitative} where in none of our solutions there is an overlap.
    \item \textbf{w/ rotation-inv.}:  We replace the rotation-equivariant backbone with a rotation-invariant one, removing the inductive bias toward learning orientation-consistent features. Rotation RMSE increases by 14.1\%, translation RMSE rises by 48.9\%, PA decreases by 9.5 pp, and CD worsens by 71.0\%. 
    These results show that rotation-invariance degrades performance, as it removes orientation cues essential for alignment and thus the model cannot distinguish rotated inputs, resulting in larger errors.
    \item \textbf{w/o rotation-equiv.:} We replace the rotation-equivariant backbone with a standard architecture lacking any inductive bias toward orientation consistency. CD increases most (29\%), while rotation error increases by 12\%, PA decreases by 3.6 pp, and translation error increases by 5\%. 
    These results highlight that rotation-equivariant features act as a crucial inductive bias for this task, enabling consistent reasoning under arbitrary orientations and improving alignment stability and generalization.

\end{itemize}


\begin{table}[htb!]
\centering
\setlength{\tabcolsep}{3pt} 
\resizebox{0.95\columnwidth}{!}{%
\begin{tabular}{@{} l r r r r @{}}
\toprule
\textbf{Methods}
  & \multicolumn{1}{c}{\textbf{RMSE ($\mathcal{R}^\circ$)↓}} 
  & \multicolumn{1}{c}{\textbf{RMSE ($\mathcal{T}_{mm}$)↓}} 
  & \multicolumn{1}{c}{\textbf{PA (\%)$\uparrow$ }}
  & \multicolumn{1}{c}{\textbf{CD}↓} \\ 
\midrule
\multicolumn{5}{c}{\textit{Tested on the Everyday Subset}} \\
\midrule
Global~\cite{li2020learning} & 80.50 & 14.60 & 28.70 & 13.00 \\
LSTM~\cite{wu2020pq} & 82.70 & 15.10 & 27.50 & 13.30 \\
DGL~\cite{zhan2020generative} & 80.30 & 13.90 & 31.60  & 11.80 \\
Jigsaw~\cite{Lu2023JigsawLT}     & 42.19 & 6.85  & 68.89 & 8.22  \\
PMTR~\cite{Lee20243DGS}       & 31.57 & 9.95  & 70.60 & 5.56  \\
PF++~\cite{Wang2024PuzzleFusionA3}       & 35.61 & 6.05  & 76.17 & 2.78  \\
GARF-mini~\cite{Li2025GARFLG}  & 6.68  & 1.34  & 94.77 & 0.25  \\
GARF~\cite{Li2025GARFLG}       & \underline{6.10}  & \underline{1.22}  &\underline{95.33} & \underline{0.22}  \\
  \hdashline
\textbf{\ourmethod} (w/o color)   & \textbf{5.31}  & \textbf{1.14}  & \textbf{96.20} & \textbf{0.18}  \\
\midrule
\multicolumn{5}{c}{\textit{Tested on the Artifact Subset}} \\
\midrule
Jigsaw~\cite{Lu2023JigsawLT}     & 43.75 & 7.91  & 65.12 & 8.50  \\
PF++~\cite{Wang2024PuzzleFusionA3}       & 47.03 & 10.63 & 57.97 & 8.24  \\
GARF-mini~\cite{Li2025GARFLG}  & 7.67  & 1.77  & 93.34 & 0.81  \\
GARF~\cite{Li2025GARFLG}       & \underline{5.82}  & \underline{1.27}  & \underline{95.04} & \underline{0.42}  \\
  \hdashline
\textbf{\ourmethod} (w/o color) & \textbf{4.63}  & \textbf{1.07}  & \textbf{96.80} & \textbf{0.20}  \\
\bottomrule
\end{tabular}%
} 
\caption{Quantitative results on Breaking Bad dataset.}
\label{tab:performance}
\end{table}

Table \ref{tab:performance} summarizes the results on the Breaking Bad dataset. 
Because this dataset lacks color information, we evaluate \ourmethod with the geometry stream only of the architecture. 
On the Everyday subset, our method achieves the best overall performance, reducing the rotational error by 13.0\%, the translational error by 6.6\% and the CD by 18.2\% compared to the strongest baseline. 
The PA improves by 0.87 pp.
On the Artifact subset, our method again leads on all metrics, yielding up to 16.3\% and 15.8\% lower rotational and translation error respectively and a massive 35.5\% reduction in CD while PA improves by 1.56 pp. 
On average across the two subsets, the improvements correspond to roughly 15\% lower rotation error, 11\% lower translation error, 27\% lower CD, and 1.2 pp PA compared to the second best competitor model, GARF.

\subsection{Ablation Studies}
\label{subsec:ablation}

We assess \ourmethod’s zero-shot generalization on Presious (color+geometry) and Fantastic Breaks (geometry-only), and fine-tuning from RePAIR and Breaking Bad for the respective scenarios.

Table~\ref{tab:3d_results_fantastic_breaks} shows the results on Fantastic Breaks a real-scan, geometry-only benchmark, indicating that \ourmethod's equivariant geometry backbone successfully generalizes beyond the training distribution. 
Despite the lack of color, our method outperforms~\cite{Li2025GARFLG} in rotation and translation by 25.0\%, while being on par for the other metrics.

\begin{table}[htb!]

\centering
\setlength{\tabcolsep}{3pt} 
\resizebox{0.95\columnwidth}{!}{%
\caption{{Quantitative results on Fantastic Breaks dataset}. 
This includes manually collected real-world objects.}
\label{tab:3d_results_fantastic_breaks}
\begin{tabular}{lcccc}
\toprule
\textbf{Methods}
  & \multicolumn{1}{c}{\textbf{RMSE ($\mathcal{R}^\circ$)↓}} 
  & \multicolumn{1}{c}{\textbf{RMSE ($\mathcal{T}_{mm}$)↓}} 
  & \multicolumn{1}{c}{\textbf{PA (\%)$\uparrow$ }}
  & \multicolumn{1}{c}{\textbf{CD}↓} \\
\midrule
Jigsaw~\cite{Lu2023JigsawLT} & 26.30 & 6.43 & 73.64 & 10.47 \\
PF++~\cite{Wang2024PuzzleFusionA3} & 20.68 & 4.37 & 83.33 & 6.68 \\
GARF~\cite{Li2025GARFLG} & \underline{10.62} & \underline{2.10} & \underline{91.00} & \textbf{2.12} \\
  \hdashline
\textbf{\ourmethod} (w/o color) & \textbf{7.96} & \textbf{2.02} & \textbf{92.01} & \underline{2.19} \\
\bottomrule
\end{tabular}}
\end{table}

Without any training or fine-tuning on Presious, Table~\ref{tab:3d_results_presious} shows that \ourmethod preserves the same trends observed in Table~\ref{tab:3d_results_repair}. 
The full model consistently outperforms across all metrics except the Chamfer Distance where it follows by close, while again each ablation results in the expected performance degradation.
Removing color (w/o color) lowers PA and raises CD (appearance cues no longer help disambiguate matches);
removing the non-overlap loss (w/o non-overlapp. loss) increases CD by allowing overlapping even when poses look plausible;
replacing the rotation-equivariant backbone with a rotation-invariant one (w/ rotation-inv) negatively impacts both rotation and translation performance;
removing rotation equivariance (w/ rotation-equiv) primarily hurts rotation accuracy and slightly destabilizes translation.
Geometry-only baselines such as GARF trail the full \ourmethod, that our color with an equivariant backbone fusion transfers reliably to previously unseen data without any dataset-specific training, evidencing strong out-of-distribution generalization, see Fig.~\ref{fig:qualitative}.

\begin{table}[htb!]
\centering
\setlength{\tabcolsep}{2pt} 
\resizebox{\columnwidth}{!}{%
\begin{tabular}{lrrrr}  
 \toprule
  Method & \multicolumn{1}{c}{\textbf{RMSE ($\mathcal{R}^\circ$)↓}} 
  & \multicolumn{1}{c}{\textbf{RMSE ($\mathcal{T}_{mm}$)↓}} 
  & \multicolumn{1}{c}{\textbf{PA (\%)$\uparrow$ }}
  & \multicolumn{1}{c}{\textbf{CD}↓} \\

    \midrule
    
    DiffAssemble~\cite{scarpellini2024diffassemble} & 56.65 & 176.41 & 24.72 & 37.73\\
    PMTR~\cite{Lee20243DGS} & 80.01 & 68.33 & 14.27 & \underline{18.21}\\
    PF++~\cite{Wang2024PuzzleFusionA3} & 70.61 & 146.84 & 24.79 & 41.58\\
    GARF~\cite{Li2025GARFLG} & \underline{43.33} & \underline{14.24} & \underline{50.27} & \textbf{18.01}\\
    \textbf{\ourmethod} (w/o color) & 35.12 & 13.85 & 52.91 & 21.18\\ \hline
    \textbf{\ourmethod} (w/o non-overlapp.) loss & 36.12 & 11.91 & 50.05 & 21.43\\
      \textbf{\ourmethod} (w/ rotation-inv.) & 41.26 & 45.01 & 47.12 & 24.62  \\
    \textbf{\ourmethod} (w/o rotation-equiv.) & 39.67 & 15.72 & 46.11 & 23.71\\
    \hdashline
    \textbf{\ourmethod} & \textbf{30.26} & \textbf{12.85} & \textbf{57.49} & 20.95\\
\bottomrule
\end{tabular}}
\caption{Quantitative results on 3D puzzle solving generalization on Presious dataset (model trained on RePAIR).} 
\label{tab:3d_results_presious}
\vspace{-10pt}
\end{table}


\section{Conclusion}
\label{sec:conlcusion}


We introduce \ourmethod, a multimodal 3D reassembly framework that fuses fracture-aware geometry via an $SO(3)$-equivariant backbone with color cues, and enforces physical plausibility through a non-overlap loss. 
Evaluated on two real-world fresco collections with colored fractures and two synthetic, without color benchmarks, \ourmethod consistently outperforms all competing methods. 
These results highlight the potential of \ourmethod for robust, generalizable 3D reconstruction in both synthetic and real-world settings.

\noindent\textbf{Limitations \& Future Work.} Point count per object affects accuracy (see suppl. material), but higher counts raise compute and memory cost. 
We will prioritize scalability via linearized attention and other efficiency-oriented architectures. 
A further limitation is our point-cloud input, which can miss high-resolution geometric and appearance detail; to address this, we plan a mesh-first variant using triangle meshes and texture maps to exploit connectivity and per-face cues.

\section*{Acknowledgements}
This work is part of the RePAIR project that has received funding from the European Union’s Horizon 2020 research and innovation programme under the grant agreement No.~964854.

{
    \small
    \bibliographystyle{ieeenat_fullname}
    \bibliography{main}
}


 \clearpage
 \setcounter{page}{1}
 \maketitlesupplementary
 \appendix

 This document expands the main paper along three areas.

\textbf{Section~\ref{app:foundations_of_symmetry} (Groups, Symmetries and Equivariance).} We formalize the symmetry principles behind our model where a fragment collection is treated as an \emph{unordered set} acted on by the symmetric group \(S_N\), meanwhile geometry lives in \(\mathbb{R}^3\) under rigid motions \(SE(3)\). We justify that the learned tokenization and message passing are \emph{permutation–equivariant} (\(f(\pi X)=\pi f(X)\) for any \(\pi\in S_N\)) and that feature fields are \emph{rotation–equivariant} (\(f(RX)=R\,f(X)\) for \(R\in SO(3)\)), while scalar heads remain invariant. Short derivations, notation, and sanity checks are provided.

\textbf{Section~\ref{supp:ablation} (Implementation \& ablations).} 
We add training and evaluation details and report extended ablations, including points-per-fragment, memory/compute trade-offs, fracture-surface thresholds, and the effect of the non-overlap weight.

\textbf{Section~\ref{supp:visual}  (Additional qualitative results).} We present more visual assemblies, side-by-side comparisons to the other baselines, and representative failure cases, complementing the quantitative and qualitative results presented in the main paper.

 \section{Groups, Symmetries and Equivariance}~\label{app:foundations_of_symmetry}
In this section we recall the basic notions linking group theory, symmetries, and equivariance. Abstract group theory (Sec. \ref{subsec:groups}) provides an abstract framework for describing symmetries as transformations that leave certain structures invariant. Group \textit{representations} (Sec. \ref{subsec:repr}) realize these abstract symmetries as linear transformations on vector spaces, enabling their analysis via linear algebra. Building on these concepts, equivariance (Sec. \ref{subsec:equiv}) characterizes maps that are compatible with a given group action, in the sense that applying a symmetry before or after the map yields consistent results under some group action.

 \subsection{Groups}
 \label{subsec:groups}

 \begin{definition}[Group]\label{def:group}
 A \emph{group} is a pair $(G, \cdot)$ consisting of a set $G$ together with a binary operation $\cdot : G \times G \to G$ satisfying the following axioms:
     \begin{itemize}
         \item[\textit{(A1)}] \textit{Associativity:} $(a \cdot b) \cdot c = a \cdot (b \cdot c)$ for all $a,b,c \in G$;
         \item[\textit{(A2)}] \textit{Identity:} there exists an element $e \in G$ such that $e \cdot a = a \cdot e = a$ for all $a \in G$;
         \item[\textit{(A3)}] \textit{Inverses:} for each $a \in G$, there exists an element $a^{-1} \in G$ satisfying $a^{-1} \cdot a = a \cdot a^{-1} = e$.
     \end{itemize}
 We typically write the operation and omit the symbol ``$\cdot$’’ when unambiguous. If, in addition, the binary operation $\cdot$ is commutative, that is, if $a \cdot b = b \cdot a$ for all $a,b \in G$, we say the group is \emph{abelian}.
 \end{definition}

We present the following groups, which formalize discrete and continuous symmetries of interest for the fragments introduced in the main paper (Sec. \ref{subsec:encoding-fragments}):

\begin{definition}[Symmetric group $S_N$] \label{def:sym_group}
For a finite set $[N] = {1, \dots, N}$, the set of all bijections $\sigma : [N] \to [N]$ forms the \emph{symmetric group} on $N$ elements,
\[
S_N = \{\; \sigma: [N] \to [N] \; | \; \sigma \text{ is bijective } \;\},
\]
under composition of maps. The identity element is the identity map $\mathrm{id}$.
\end{definition}

This notion extends naturally from discrete sets to linear spaces.

\begin{definition}[General Linear Group GL(V)]\label{def:gl}
Let $V$ be a finite-dimensional real vector space. The \emph{general linear group} of $V$ is
\[
GL(V) = \{\, T : V \to V \mid T \text{ is linear and invertible} \,\},
\]
with composition as the group operation. Choosing a basis of $V$ identifies each $T\in GL(V)$ with an invertible matrix, yielding the matrix group
\[
GL_N(\mathbb{R}) = \{\, A \in \mathbb{R}^{N\times N} \mid \det(A)\neq 0 \,\},
\]
where $N=\dim(V)$.
\end{definition}

\begin{definition}[Special Orthogonal Group $SO(V)$]\label{def:so_abstract}
Let $V$ be an $N$-dimensional real inner-product space with inner product $\langle \cdot,\cdot \rangle$. The \emph{special orthogonal group} of $V$ is
\begin{align*}
SO(V)
= \{\; T \in GL(V)\ \mid\ &
\langle Tv, Tw\rangle = \langle v, w\rangle \\
\text{ for all } v,w \in V,
& \det(T)=1 \;\}.
\end{align*}
Its elements are the orientation-preserving linear isometries of $V$.
\end{definition}

This definition has no coordinates in it; the structure depends only on the inner product and the notion of orientation. A basis simply provides a concrete representation.

\begin{observation}
Choosing an orthonormal basis of $V$ identifies $SO(V)$ with the matrix group
\[
SO(N) = \{\, R \in \mathbb{R}^{N\times N} \mid R^\top R = I,\; \det(R)=1 \,\},
\]
since linear isometries correspond exactly to matrices preserving the standard inner product.
\end{observation}

Specializing to three dimensions gives the familiar rotation group in Euclidean space.

\begin{definition}[Special Orthogonal Group SO(3)]\label{def:so3_matrix}
The \emph{special orthogonal group} in three dimensions is
\[
SO(3) = \{\, R \in \mathbb{R}^{3\times 3} \mid R^\top R = I,\; \det(R)=1 \,\},
\]
the group of proper rotations of $\mathbb{R}^3$.
\end{definition}

\begin{definition}[Special Euclidean Group SE(V)]\label{def:se_abstract}
Let $V$ be an $N$-dimensional real inner-product space. The \emph{special Euclidean group} of $V$ is the group of orientation-preserving isometries $f:V\to V$. Each such isometry has a unique decomposition 
\[
f(x)=T x + t, \qquad T\in SO(V),\; t\in V,
\]
and composition is given by
\[
(T_1,t_1)(T_2,t_2) = (T_1 T_2,\; T_1 t_2 + t_1).
\]
\end{definition}

\begin{definition}[Special Euclidean Group SE(3)]\label{def:se3}
Specializing to $V=\mathbb{R}^3$ under a basis, one obtains
\[
SE(3)=\{(R,t)\mid R\in SO(3),\; t\in\mathbb{R}^3\},
\]
acting on $x\in\mathbb{R}^3$ by $x\mapsto R x+t$, with the composition law
\[
(R_1,t_1)(R_2,t_2) = (R_1R_2,\; R_1 t_2 + t_1).
\]
\end{definition}

\begin{observation}[Semidirect-product form]
The map
\[
SO(3)\ltimes \mathbb{R}^3 \;\longrightarrow\; SE(3),\qquad (R,t)\mapsto (x\mapsto R x + t),
\]
is a group isomorphism. Thus $SE(3)$ is naturally identified with the semidirect product $SO(3)\ltimes \mathbb{R}^3$, where $SO(3)$ acts on $\mathbb{R}^3$ by its standard linear action.
\end{observation}

\subsection{Representations}
\label{subsec:repr}
Representations make concrete the notion of symmetry by describing how elements of a group act linearly on a vector space. Each group element is associated with an invertible linear transformation, transferring the group’s algebraic structure into linear operators.

\begin{definition}[Linear Group Action]\label{def:repr}
Let $G$ be a group and $V$ a finite-dimensional real vector space. A \emph{representation} of $G$ on $V$ is a group homomorphism
\[
\rho_V : G \to GL(V),
\]
which defines the action
\[
g \cdot v := \rho_V(g)\,v, \qquad g\in G,\; v\in V.
\]
\end{definition}

\begin{definition}[Linear Group Action]\label{def:repr}
Let $V$ be a finite-dimensional vector space, and let $G$ be a finite group. A representation $V$ or linear group action is a group homomorphism
\[
\rho^V: G \to GL(V),
\]
which defines the action 
\[
\rho^V(g) \cdot v  \quad \text{ for all } \quad g \in G, v \in V.
\]
\end{definition}

 The symmetric group (Definition~\ref{def:sym_group}) admits a representation permuting coordinate entries.

\begin{definition}[Standard Representation of $S_N$]\label{def:standard_repr}
Let $V = \mathbb{R}^N$ with canonical basis $\{e_i\}_{i=1}^N$. The \emph{standard representation} of the symmetric group $S_N$ acts by permuting coordinates:
\[
(\sigma \cdot x)_i = x_{\sigma^{-1}(i)}, \qquad \forall\, \sigma \in S_N, \, x = [x_1,\ldots,x_N] \in \mathbb{R}^N.
\]
Equivalently, this action can be expressed as left multiplication by the permutation matrix $P_{\sigma} \in \mathrm{GL}_N(\mathbb{R})$, defined by
\[
(P_{\sigma})_{ij} =
\begin{cases}
1, & \text{if } i = \sigma(j),\\
0, & \text{otherwise.}
\end{cases}
\]
This realizes the standard representation as a concrete matrix representation of $S_N$.
\end{definition}

Continuous groups admit analogous linear actions.
For $V=\mathbb{R}^3$, the inclusion $SO(3) \subset GL_3(\mathbb{R})$ defines the standard representation 
\[
R \cdot x = R x, \qquad R \in SO(3),\; x\in\mathbb{R}^3,
\]
given by left multiplication by rotation matrices.

\subsection{Equivariance and Invariance}
\label{subsec:equiv}

Equivariance and invariance describe how maps between representations interact with group actions. They formalize when a transformation “respects’’ symmetry. Such description in exploited in section \ref{subsec:encoding-fragments} where the Geometric and Color encoders are introduced.

\begin{definition}[Equivariant Map]\label{def:equivariant}
Let $(V,\rho_V)$ and $(W,\rho_W)$ be representations of a group $G$.
A (not necessarily linear) map $f : V \to W$ is \emph{G-equivariant} if
\[
f(\rho_V(g)\,v) = \rho_W(g)\,f(v) \qquad \text{for all } g\in G,\; v\in V.
\]
Equivariance means that applying a group action before or after $f$ yields the same outcome.
\end{definition}

\begin{definition}[Invariant Map]\label{def:invariant}
Let $(V,\rho_V)$ and $(W,\rho_W)$ be representations of a group G.
A (not necessarily linear) map $f : V \to W$ is \emph{G-invariant} if
\[
f(\rho_V(g)\,v) = f(v) \qquad \text{for all } g\in G,\; v\in V.
\]
Invariance means that f is unaffected by the action of G on its input.
\end{definition}

\section{Additional Info and Ablation Studies}\label{supp:ablation}

Here we present additional details related to \ourmethod's training and evaluation pipelines (Section~\ref{supp:additional_info}) and extended ablations on the key factors that drive accuracy and cost (Section~\ref{supp:ablation_studies}). 
We first study the number of points sampling, showing how increasing point count improves performance while exhibiting diminishing returns. 
We then quantify memory usage as point density rises. 
Next, we analyze fracture-surface ground-truth sensitivity on RePAIR by varying the distance threshold used in pretraining, which is important given the dataset’s random erosion, and report its effect on downstream accuracy. 
Finally, we sweep the no-overlap loss weight $\alpha$ to characterize the trade-off between collision suppression~(CD/intersections) and part accuracy~(PA), selecting a stable default.

\subsection{Additional Info}\label{supp:additional_info}
\noindent\textbf{Competing Methods.} We compare against GARF~\cite{Li2025GARFLG}, DiffAssemble~\cite{scarpellini2024diffassemble}, PMTR~\cite{Lee20243DGS}, PuzzleFusion++~(PF++)~\cite{Wang2024PuzzleFusionA3} and Jigsaw~\cite{Lu2023JigsawLT} as the most recent and representative state-of-the-art for multi-fragment 3D assembly on point clouds, and they collectively span the main methodological families: \textit{(i)} generative diffusion/flow pipelines for pose refinement (DiffAssemble), \textit{(ii)} matching-based pose estimation with efficient point-cloud correspondences (PMTR), \textit{(iii)} search/verify agglomerative assembly (PF++ and Jigsaw), and (iv) a generalization-oriented geometry SOTA with strong results on real fractures (GARF).

\subsection{Ablation Studies}\label{supp:ablation_studies}
\subsubsection{Number of Keypoints \& Memory}

We compare \ourmethod with GARF~\cite{Li2025GARFLG} across different numbers of points on the RePAIR dataset (Table~\ref{tab:ablation_repair}). 
\ourmethod{} consistently outperforms GARF across all metrics, achieving lower rotation and translation errors, higher part accuracy, and reduced Chamfer Distance.

Notably, \ourmethod{} exhibits a positive correlation between the number of points and performance, highlighting our method's ability to leverage denser point clouds for more accurate fragment alignment.
%
In contrast, this trend is not observed in GARF, where adding more points does not necessarily aid its reconstruction capabilities. We attribute this behavior to the fact that GARF does not utilize color information. By effectively leveraging the richer color signals present in denser point clouds, \ourmethod{} maximizes reconstruction accuracy.

\begin{figure}[htb!]
    \centering
    \begin{minipage}{1\linewidth}
        \begin{tikzpicture}
        \begin{axis}[
            height=4.5cm,
            width=\linewidth,
            xlabel={Number of sampled points},
            ylabel={Memory (GB)},
            xmin=0.5, xmax=4.5,
            ymin=0, ymax=50,
            ytick={10,20,30,40,50},
            xtick={1,2,3,4},
            xticklabels={5K, 10K, 15K, 20K},
            xmajorgrids=false,
            ymajorgrids=true,
            grid style={line width=.1pt, draw=gray!10},
            major grid style={line width=.2pt,draw=gray!50},
            legend style={at={(0.5, 1.0)}, anchor=south, legend columns=2},
            axis lines=left,
            axis on top,
            clip=false
        ]

        \addlegendentry{\footnotesize{\ourmethod{}}}
        \addplot[color=col1, mark=square, line width=1pt] coordinates {
            (1, 6.64)
            (2, 11.74)
            (3, 17.02)
            (4, 22.17)
        };
        \addlegendentry{\footnotesize{GARF}}
        \addplot[color=blue, mark=o, line width=1pt] coordinates {
            (1, 5.38)
            (2, 9.16)
            (3, 13.00)
            (4, 17.10)

        };

        \addlegendentry{\footnotesize{Nvidia A100 40GB}}
        \addplot[color=red, dashed, line width=2pt] coordinates {
            (0.5, 40)
            (4.5, 40)
        };

        \end{axis}
        \end{tikzpicture}
        \caption{\label{fig:memory}GPU memory consumption on the RePAIR dataset as a function of the varying number of sampled points (5K–20K).}
    \end{minipage}
\end{figure}
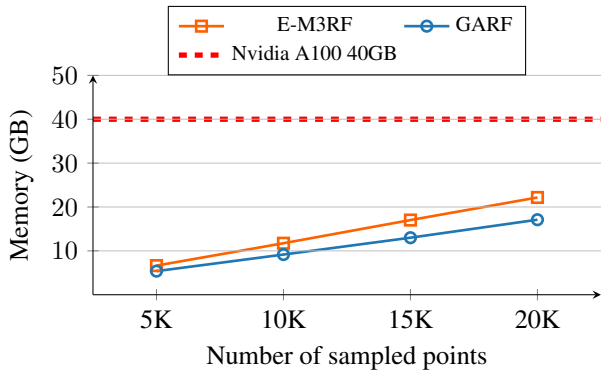

\paragraph{Computational Cost.} It is important to note that higher point densities naturally demand greater computational resources. As shown in Figure~\ref{fig:memory}, memory usage for both models increases approximately linearly with the number of points. Although \ourmethod{} exhibits slightly higher memory consumption than GARF, primarily due to the additional color features, the peak usage remains comfortably within our hardware limits (NVIDIA A100, 40GB).

\begin{table}[htb!]
\centering
\setlength{\tabcolsep}{5pt}
\resizebox{\columnwidth}{!}{%
\begin{tabular}{l lrrrrr}
 \toprule
 \#Points & Method & 
 \multicolumn{1}{c}{\textbf{RMSE ($\mathcal{R}^\circ$)↓}} &
 \multicolumn{1}{c}{\textbf{RMSE ($\mathcal{T}_{mm}$)↓}} &
 \multicolumn{1}{c}{\textbf{PA (\%)$\uparrow$}} &
 \multicolumn{1}{c}{\textbf{CD}↓} \\
 \midrule

 \multirow{2}{*}{5K} 
   & GARF~\cite{Li2025GARFLG} & 49.31 & 32.19 & 31.14 & {2.66}\\
   & \textbf{\ourmethod} & 37.91 & 27.93 & 35.91 &  2.17\\[2pt]

 \multirow{2}{*}{10K} 
   & GARF~\cite{Li2025GARFLG} & \underline{43.92} & 30.12 & 32.55 & \underline{2.52} \\
   & \textbf{\ourmethod} & 35.82 & 26.01 & {37.15} &  2.12\\[2pt]

 \multirow{2}{*}{15K} 
   & GARF~\cite{Li2025GARFLG} & 48.27 & 28.02 & 33.10 & 3.12 \\
   & \textbf{\ourmethod} & 37.21 & 26.04 & 36.51 &  \textbf{2.10}\\[2pt]

 \multirow{2}{*}{20K} 
   & GARF~\cite{Li2025GARFLG} & 45.56 & \underline{27.94} & \underline{34.21} &  3.01\\
   & \textbf{\ourmethod} & \textbf{34.92} & \textbf{25.03} & \textbf{37.21} &  2.11\\

 \bottomrule
                                                      \end{tabular}}
\caption{Quantitative results on the RePAIR dataset with varying number of points (5K–20K). \textbf{Bold} indicates the best result, while \underline{underlined} values denote the competitor best result. }
\label{tab:ablation_repair}

\end{table}

\subsubsection{Distance Threshold for Fracture-surface}


{The Fracture-Boundary Segmentation Pretraining task is introduced as a large-scale pretraining approach to learn fragment representations to effectively capture the geometric signatures of the fracture interfaces, crucial for accurate reassembly.
The ground truth label for this task is derived by leveraging a distance threshold $\tau$ to identify contact points between two fragments, where $\tau$ controls how close points need to be to count as contact.}

For the RePAIR dataset experiments reported in the main paper, we set the distance threshold to $0.4$. This value provides a balance between robustness to gaps caused by random erosion or scanning noise and precision in detecting true contacts. We note that smaller thresholds risk miss valid connections, while larger thresholds can falsely connect fragments that are not actually adjacent.
In Table~\ref{tab:ablation_distance_thresholds}, we present an ablation study varying the distance threshold to analyze its effect on fragment connectivity and reconstruction. This experiment demonstrates how the choice of threshold impacts the detection of shared surfaces and the resulting reassembly.

\begin{table}[htb!]
\centering
\setlength{\tabcolsep}{4pt} 
\resizebox{0.95\columnwidth}{!}{%
\caption{Ablation study on fracture surface ground-truth generation for RePAIR~\cite{Tsesmelis2024ReassemblingTP}. 
Different distance thresholds for defining fracture-surface ground truth are evaluated.}
\label{tab:ablation_distance_thresholds}
\begin{tabular}{lcccc}
\toprule
\textbf{$\tau$} 
  & \multicolumn{1}{c}{\textbf{RMSE ($\mathcal{R}^\circ$)↓}} 
  & \multicolumn{1}{c}{\textbf{RMSE ($\mathcal{T}_{mm}$)↓}} 
  & \multicolumn{1}{c}{\textbf{PA (\%)$\uparrow$}} 
  & \multicolumn{1}{c}{\textbf{CD↓}} \\
\midrule
0.2   & \underline{40.15} & 57.22 & 27.02 & 4.14 \\
0.4   & \textbf{37.91} & \underline{27.93} & \textbf{35.91} &  \textbf{2.17} \\
0.6   & 41.02 & \textbf{27.42} & \underline{33.67} & \underline{2.18} \\
0.8   & 42.71 & 34.31 & 31.29 & 3.01\\
\bottomrule
\end{tabular}}
\end{table}


\subsubsection{Alpha for No-Overlap Loss}

In Table~\ref{tab:ablation_no_overlap}, we analyze the effect of the weighting factor~$\alpha$, which balances assembly accuracy with overlap prevention. When~$\alpha$ is too small, the network produces highly accurate assemblies but occasionally predicts overlapping parts. Conversely, a large~$\alpha$ enforces physical consistency more strictly but can reduce assembly precision. Our experiments show that an intermediate value, specifically $\alpha= 0.3$, achieves the best trade-off, yielding $SE(3)$ predictions that are both accurate and physically feasible, as clearly demonstrated by the results in Table~\ref{tab:ablation_no_overlap}. This highlights the importance of the no-overlap loss term in achieving physically consistent training.
\begin{table}[htb!]
\centering
\setlength{\tabcolsep}{4pt} 
\resizebox{0.95\columnwidth}{!}{%
\caption{Ablation study on the effect of the no-overlap loss weight~($\alpha$) on the RePAIR~\cite{Tsesmelis2024ReassemblingTP} dataset.}
\label{tab:ablation_no_overlap}
\begin{tabular}{lcccc}
\toprule
\textbf{$\alpha$} 
  & \multicolumn{1}{c}{\textbf{RMSE ($\mathcal{R}^\circ$)↓}} 
  & \multicolumn{1}{c}{\textbf{RMSE ($\mathcal{T}_{mm}$)↓}} 
  & \multicolumn{1}{c}{\textbf{PA (\%)$\uparrow$}} 
  & \multicolumn{1}{c}{\textbf{CD↓}} \\
\midrule

0.1   &  43.26 & \textbf{25.92} & 33.07 &  3.42 \\
0.3   &  \textbf{37.91} & \underline{27.93} & \textbf{35.91} &  \textbf{2.17} \\
0.5   &  \underline{37.98} & 28.22 & \underline{35.02} & \underline{3.01} \\ 
0.7   &  40.12 & 33.75 & 31.05 & 4.21 \\ 
\bottomrule
\end{tabular}}
\end{table}

\section{Additional Visualizations}\label{supp:visual}

This section presents qualitative results corresponding to the four evaluation datasets—RePAIR, Presious, Breaking Bad, and Fantastic Breaks—and illustrate how~\ourmethod behaves under different appearance and fracture variations and across varying piece counts (note that Fantastic Breaks contains only 2-piece puzzles by design).

\begin{figure*}[t]
\centering
\includegraphics[width=1.\linewidth]{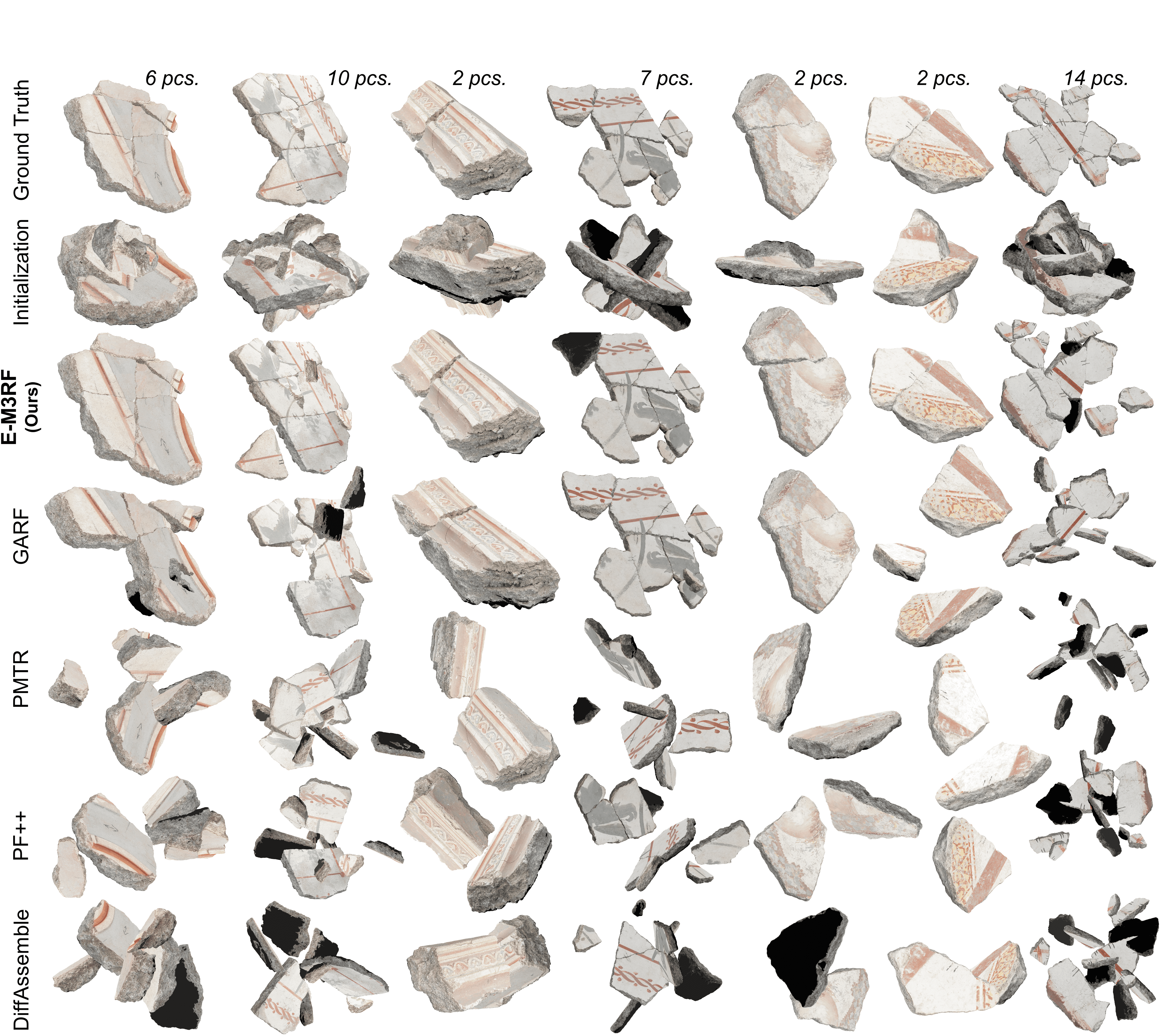}
\caption{
Qualitative Comparisons on the RePAIR. \ourmethod consistently produces more accurate re-assemblies.
} 
\label{fig:qualitative_repair}
\vspace{-1.0em}
\end{figure*}

On the RePAIR dataset, Fig.~\ref{fig:qualitative_repair}, examples with several pieces show that our multimodal fusion aligns both fracture boundaries and colored motifs. 
On the other hand competing methods often leave slight pose drift or texture misalignment on thin or eroded edges.

\begin{figure*}[t]
\centering
\includegraphics[width=.9\linewidth]{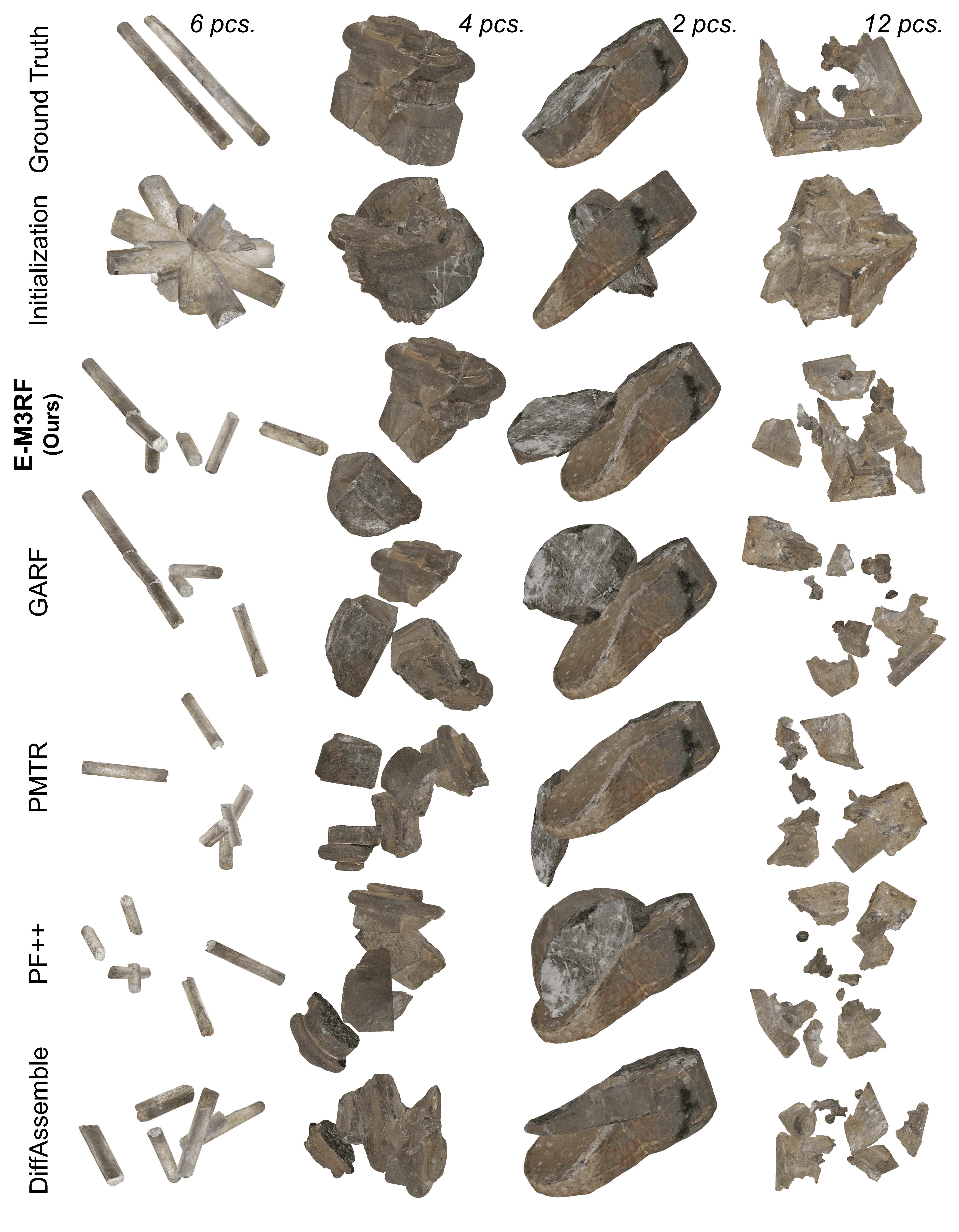}
\caption{
Qualitative Comparisons on the remaining four puzzles of the Presious dataset. 
} 
\label{fig:qualitative_presious}
\vspace{-1.0em}
\end{figure*}

On the Presious dataset, Fig.~\ref{fig:qualitative_presious}, despite the surface wear and uneven point density, our reconstructions minimize interpenetration and recover consistent contact along long, low-curvature breaks; baselines tend to over- or under-insert pieces, leaving visible gaps or overlaps.

\begin{figure*}[t]
\centering
\includegraphics[width=1.\linewidth]{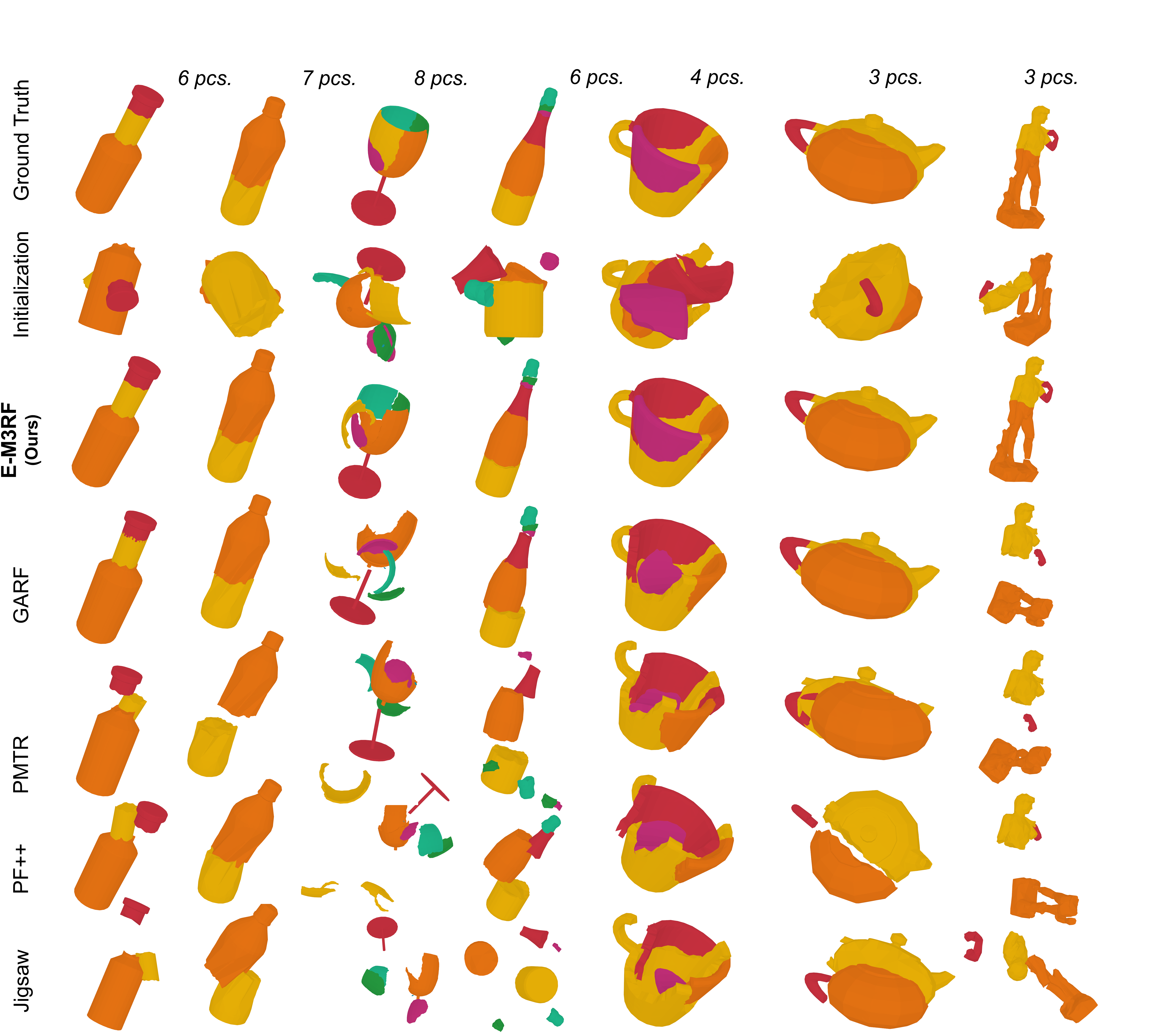}
\caption{
Qualitative Comparisons on the Breaking Bad (BB). \ourmethod consistently produces more accurate re-assemblies.
} 
\label{fig:qualitative_bb}
\vspace{-1.0em}
\end{figure*}

On the only geometry datasets and specifically on Breaking Bad, Fig.~\ref{fig:qualitative_bb}, With complex, multi-piece assemblies, the equivariant backbone maintains stable orientations and lowers rotational error, producing globally coherent placements where alternatives seem to get trapped in near-symmetric configurations.

\begin{figure*}[t]
\centering
\includegraphics[width=1.\linewidth]{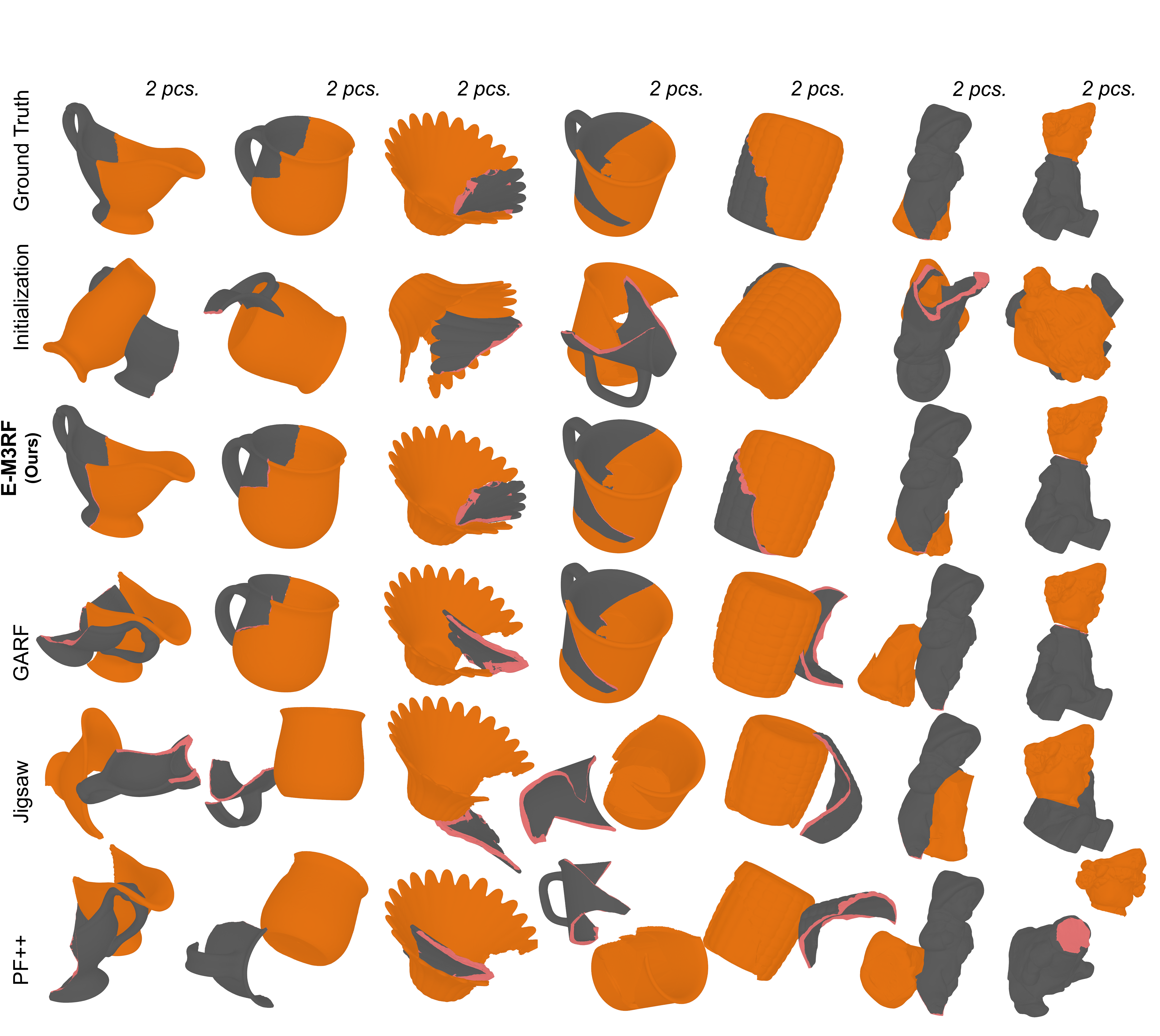}
\caption{
Qualitative Comparisons on the Fantastic Breaks. \ourmethod consistently produces more accurate re-assemblies. Note, the Fantastic Breaks dataset is composed of only 2-piece puzzles by design. 
}
\label{fig:qualitative_fb}
\vspace{-1.0em}
\end{figure*}

Finally, on the Fantastic Breaks (2-piece only, and real only geometry related scans), Fig.~\ref{fig:qualitative_fb}, it seems that our no-overlap loss prevents subtle collisions while achieving tight, flush joins; while others show slight interpenetration or residual offsets.

Overall, the visuals echo the quantitative results:~\ourmethod consistently achieves tighter pose alignment, fewer collisions, and—when the color cues are available—better cross-shard pattern continuity, while remaining robust on geometry-only data.
Figures show qualitative comparisons of \ourmethod{} against GARF~\cite{Li2025GARFLG}, DiffAssemble~\cite{scarpellini2024diffassemble}, PMTR~\cite{Lee20243DGS}, PF++~\cite{Wang2024PuzzleFusionA3} and Jigsaw~\cite{Lu2023JigsawLT} on the four benchmarks. 
%
 
\end{document}

%% file: preamble.tex
\usepackage{graphicx}
\usepackage{amsmath}
\usepackage{amssymb}
\usepackage{booktabs}
\usepackage{subfiles}
\graphicspath{{imgs/}{../imgs/}}

\usepackage{bbm}

\usepackage{url}

\usepackage{float}
\floatstyle{plaintop}
\restylefloat{table}

\usepackage{esvect}
\usepackage{booktabs,makecell}
\usepackage{colortbl}
\usepackage{xcolor}
\usepackage{wrapfig}
\usepackage{float}
\restylefloat{table}
\usepackage{placeins}
\usepackage{stfloats}
\usepackage{multicol}
\usepackage{tabu,multirow}
\usepackage{tabularx}
\usepackage{nicematrix}
\usepackage{booktabs}
\usepackage{placeins}
\usepackage{booktabs}
\usepackage{footnote}
\usepackage{afterpage}
\usepackage{adjustbox}
\usepackage{arydshln}
\usepackage{gensymb}
\usepackage{caption}
\usepackage{subcaption}
\usepackage{tabularray}

\usepackage{mathtools}  

\usepackage{siunitx} 

\usepackage{tikz} 

\usepackage{pifont}

\definecolor{lightgrey}{rgb}{0.7, 0.7, 0.7} 

\newcolumntype{"}{@{\hskip\tabcolsep\vrule width 1.3pt\hskip\tabcolsep}}

\newcolumntype{P}[1]{>{\centering\arraybackslash}m{#1}}

\let\svthefootnote\thefootnote
\newcommand\freefootnote[1]{%
  \let\thefootnote\relax%
  \footnotetext{#1}%
  \let\thefootnote\svthefootnote%
}

\newcommand{\encircle}[2][]{%
  \tikz[baseline=(char.base)]{%
    \node[shape = circle, draw, inner sep = 1pt]
    (char) {\phantom{\ifblank{#1}{#2}{#1}}};%
    \node at (char.center) {\makebox[0pt][c]{#2}};}}
\robustify{\encircle}